\documentclass{article}
\PassOptionsToPackage{numbers,sort&compress}{natbib}

\usepackage[main, final]{neurips_2026}

\usepackage[utf8]{inputenc} 
\usepackage[T1]{fontenc}    
\usepackage{hyperref}       
\usepackage{url}            
\usepackage{xurl}
\usepackage{booktabs}       
\usepackage{amsfonts}       
\usepackage{nicefrac}       
\usepackage{microtype}      
\usepackage[table]{xcolor}         
\usepackage{amsmath}
\usepackage{amssymb}
\usepackage{mathtools}
\usepackage{amsthm}
\usepackage[capitalize,noabbrev]{cleveref}
\usepackage{amssymb} 
\ifdefined\DeclareUnicodeCharacter
\DeclareUnicodeCharacter{03C0}{\ensuremath{\pi}}
\DeclareUnicodeCharacter{2192}{\ensuremath{\rightarrow}}
\fi
\usepackage{multirow}
\usepackage{array}
\usepackage[most]{tcolorbox}
\usepackage{listings}
\usepackage{listingsutf8}
\usepackage{graphicx} 
\usepackage{xspace}
\usepackage{caption}
\usepackage{wrapfig}
\usepackage{makecell}
\usepackage{adjustbox}
\usepackage{subcaption}
\usepackage{xcolor}
\usepackage{longtable}
\usepackage{needspace}

\definecolor{chgu_edit}{HTML}{C78F00}     
\definecolor{chgu_suggest}{HTML}{1E6FD9}  
\definecolor{chgu_note}{HTML}{007A33}     
\newcommand{\chgumod}[2][]{%
  {\color{chgu_edit}#2}%
  \ifx&#1&\else
    \quad{\color{chgu_note}[#1]}%
  \fi
}

\definecolor{rowgray}{gray}{0.94}
\definecolor{rowgray}{RGB}{245,245,245}
\definecolor{avgGreen}{RGB}{46,125,50}
\definecolor{taskPurple}{RGB}{123,31,162}

\newcolumntype{S}{>{\centering\arraybackslash}m{1.8cm}}

\newcommand{\avgBest}[1]{%
  \cellcolor{avgGreen!15}\textbf{#1}%
}

\newcommand{\taskBest}[1]{%
  \cellcolor{taskPurple!15}\textbf{#1}%
}

\newcommand{\avgSecond}[1]{%
  \begingroup
  \setlength{\fboxsep}{0pt}%
  \setlength{\fboxrule}{0.6pt}%
  \fcolorbox{avgGreen!55!black}{white}{%
    \makebox[\dimexpr\linewidth-1\fboxrule\relax][c]{%
      \rule[-0.55ex]{0pt}{3.0ex}#1%
    }%
  }%
  \endgroup
}

\newcommand{\taskSecond}[1]{%
  \begingroup
  \setlength{\fboxsep}{0pt}%
  \setlength{\fboxrule}{0.6pt}%
  \fcolorbox{taskPurple!85!black}{white}{%
    \makebox[\dimexpr\linewidth-1\fboxrule\relax][c]{%
      \rule[-0.55ex]{0pt}{3.0ex}#1%
    }%
  }%
  \endgroup
}

\newcolumntype{P}[1]{>{\centering\arraybackslash}p{#1}}
\usepackage{tabularx}
\lstdefinelanguage{json}{
    basicstyle=\ttfamily\small, 
    numbers=none, 
    showstringspaces=false, 
    breaklines=true, 
    morestring=[b]",
    morecomment=[l]{//},
    morecomment=[s]{/*}{*/},
    morekeywords={true,false,null}
}

\NewDocumentCommand{\heng}
{ mO{} }{\textcolor{red}{\textsuperscript{\textit{Heng}}\textsf{\textbf{\small[#1]}}}}
\definecolor{morandiGreen}{RGB}{80, 130, 100}
\definecolor{morandiYellow}{RGB}{200, 140, 80}
\definecolor{morandiBlue}{RGB}{70, 100, 140}
\definecolor{morandiPink}{RGB}{180, 80, 100}
\definecolor{lightblue}{RGB}{220,235,252}
\usepackage{pifont}
\newcommand{\cmark}{\textcolor{green!60!black}{\ding{51}}}
\newcommand{\xmark}{\textcolor{red!70!black}{\ding{55}}}

\newcommand{\geomindcolor}{%
  \textbf{%
    \textcolor[HTML]{EEBFB9}{\textbf{\textsc{Geo}}}%
    \textcolor[HTML]{9891D7}{\textbf{\textsc{Mind}}}%
  }%
}

\providecommand{\geomindbase}{\textsc{GeoMind-Base}\xspace}
\providecommand{\geomindaug}{\textsc{GeoMind-Aug}\xspace}
\providecommand{\geomindbf}{\textbf{\textsc{GeoMind}}\xspace}
\providecommand{\geomind}{\text{\textsc{GeoMind}}\xspace}

\newcommand{\sure}[1]{{\color{black}#1}}
\title{\geomindcolor: A Benchmark for Spatial Understanding in Robotic Manipulation}

\author{%
  \textbf{Jinghe Wang}\textsuperscript{1}\thanks{Equal contribution.}\quad
  \textbf{Xinrui Cao}\textsuperscript{1}\footnotemark[1]\quad
  \textbf{Duo Wu}\textsuperscript{1}\footnotemark[1]\quad
  \textbf{Chenghao Gu}\textsuperscript{1}\quad
  \textbf{Yong Zhong}\textsuperscript{2}\\[0.5ex]
  \textbf{Linjia Kang}\textsuperscript{1}\quad
  \textbf{Tianyi Xiong}\textsuperscript{1}\quad
  \textbf{Zhi Wang}\textsuperscript{1}\thanks{Corresponding author.}\\
  \textsuperscript{1}Tsinghua Shenzhen International Graduate School,
  Tsinghua University\\
  \textsuperscript{2}Southern University of Science and Technology
}

\definecolor{rebuttalpurple}{HTML}{7030A0}
\definecolor{revieworange}{HTML}{B34D00}

\begin{document}

\maketitle

\begin{abstract}

Robotic manipulation has advanced rapidly in mapping visual observations and language instructions to actions. However, reliable manipulation in open-world environments requires robust \textit{spatial understanding}: the ability to identify targets in complex layouts (\textit{spatial reasoning}) and adapt to scenes that continuously evolve through interaction (\textit{spatial memory}). Yet existing embodied benchmarks fall short in evaluating spatial understanding: they often introduce strong visual cues that bypass spatial reasoning, and focus on static scenes that offer little insight into an agent’s capacity for spatial memory. To address this gap, we introduce \geomind, a systematic benchmark for structured spatial understanding through tabletop manipulation. \geomind comprises 54 evaluated task variants organized around spatial reasoning and spatial memory. It also incorporates a scalable pipeline that automates the generation and annotation of task instances with verified targets and executable interactions. Extensive evaluations across diverse embodied agents reveal their consistent weaknesses in spatial understanding. Our findings highlight that these agents struggle to track identities after spatial changes and convert relational or geometric constraints into executable actions. We will release the \geomind benchmark at \url{https://github.com/Notfound-JH/GEOMIND-benchmark.git}.


\end{abstract}

\raggedbottom
\section{Introduction}
\begin{figure}[t]
\centering
\includegraphics[width=\linewidth]{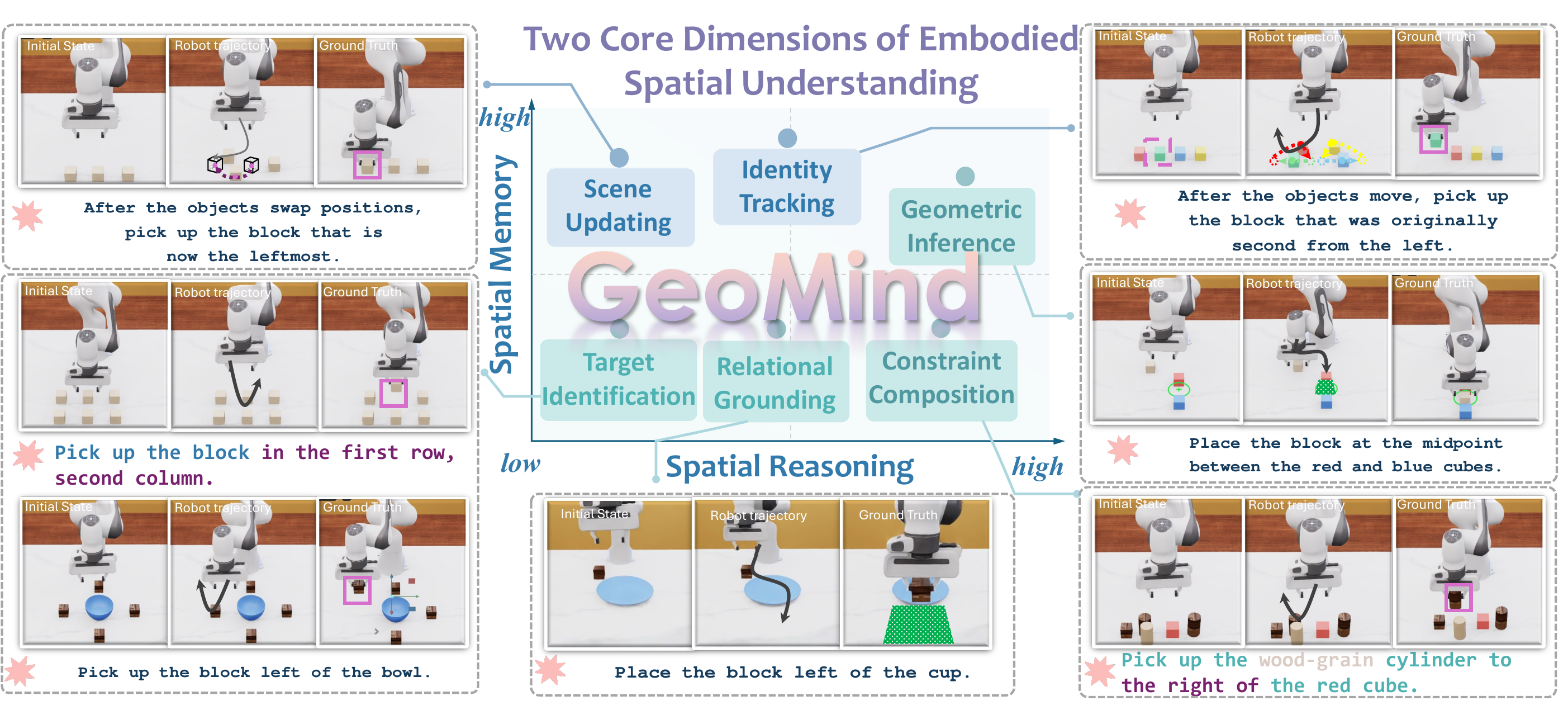}
\caption{Overview of \geomindbf. 
\geomind organizes structured spatial understanding along two diagnostic dimensions: \textit{spatial memory} and \textit{spatial reasoning}. 
The six task families cover different combinations of these dimensions. 
Surrounding examples illustrate how \geomind couples spatial instructions with structured object layouts and scene changes. The instructions are illustrative; evaluated actions and success criteria are specified in~\ref{app:task_details}.
}
\vspace{-1em}
\label{fig:cap_tasks}
\end{figure}

\sure{Advances in embodied intelligence have enabled robots to translate visual observations and language instructions into manipulation actions~\cite{gupta2021embodied,generalist2025gen0,openvla,pi0_2024,black2025pi,bu_univla_2025,liu_rdt-1b_2024,rt2,dreamzero,zerowam,openwam_wang,openwam_yu,aspire,enpire}. However, reliable manipulation often requires more than recognizing individual objects: robots must reason about spatial relationships to identify targets and maintain task-relevant spatial information as scenes change~\cite{qu_spatialvla_2025,internvlam1,zhang_spatial_2025}. For example, a robot may be asked to pick up a block based on its position relative to other objects, or to pick up the same block after observing several visually similar blocks being rearranged. While the former requires reasoning about the current spatial configuration, the latter additionally requires tracking object identity across scene changes. Such tasks highlight the importance of spatial understanding for reliable manipulation.}

\sure{Existing embodied benchmarks~\cite{liu_libero_2023,gu2023maniskill2,james2020rlbench,zhang2024vlabench,yu_meta-world_2021,jiang2023vima,chen2025robotwin} have made substantial progress in evaluating language-guided manipulation and generalization. However, overall task success provides limited insight into the spatial capabilities underlying an agent's behavior. Distinctive visual or semantic cues may allow agents to identify targets without reasoning about spatial relationships. Similarly, tasks involving scene changes may be solved from the current observation without tracking previous spatial states. Conversely, manipulation failures may arise from execution errors even when the intended spatial target has been correctly identified. These limitations motivate a diagnostic evaluation that makes spatial requirements explicit and examines agents' spatial capabilities beyond overall task success.}

\sure{To enable a more targeted assessment of spatial understanding, we focus on two key dimensions, \textbf{\textit{spatial reasoning}} and \textbf{\textit{spatial memory}}, drawing on the cognitive distinction between reasoning about space and maintaining spatial information over time~\cite{yang2024think}. Spatial reasoning concerns identifying target objects and inferring goal locations or regions from spatial relationships and structured layouts, while spatial memory concerns tracking object identities and updating spatial states as scenes change. These dimensions specify the spatial capabilities that each task is designed to assess.}

\sure{We introduce \geomindbf, a benchmark for systematically evaluating spatial understanding through tabletop robotic manipulation. It comprises six task families: (1) \textit{target identification}, (2) \textit{relational grounding}, (3) \textit{constraint composition}, and (4) \textit{geometric inference} for spatial reasoning, and (5) \textit{scene updating} and (6) \textit{identity tracking} for spatial memory. Together, these families span 54 evaluated task variants: 20 Base, 7 Semantic, 7 Visual, and 20 Execution variants. Tasks are designed around structured object layouts and controlled scene changes, with explicit spatial rules for determining target objects or goal locations. Many base tasks feature visually identical or near-identical objects to reduce reliance on appearance cues. To enable reliable evaluation of these diverse spatial objectives, we develop an agent-assisted pipeline that translates spatial constraints into executable grounding code, computes ground-truth spatial targets, and verifies the resulting task instances through simulation.}

\sure{We evaluate generalist and spatially enhanced visuomotor policies alongside modular planning methods on \geomind under a unified evaluation protocol. \geomind combines a base task suite (\geomindbase) with controlled semantic, visual, and execution augmentations (\geomindaug) to probe robustness. Beyond task success rates, we use diagnostic spatial metrics and comparisons with an oracle policy to examine spatial grounding and execution behavior. Our results show that current methods can solve some basic grounding tasks but struggle to reason about structured spatial relationships and track spatial states across scene changes. Performance is sensitive to semantic augmentations, while visual and execution perturbations have more model-dependent effects.}

\sure{The main contributions of this work are threefold: \textbf{(1) A systematic benchmark for structured spatial understanding.} We introduce \geomind, which organizes spatial understanding into six task families and provides controlled base and augmented evaluations. \textbf{(2) An agent-assisted spatial grounding and verification pipeline.} We develop a pipeline that translates spatial constraints into executable grounding programs, derives ground-truth targets, and verifies task instances through simulation. \textbf{(3) A comprehensive diagnostic evaluation.} We evaluate generalist and spatially enhanced policies alongside modular planning methods using diagnostic spatial metrics and oracle comparisons, revealing limitations in spatial reasoning, memory, and robustness.}

\section{Related Work}
\label{sec:related_work}

\sure{\textbf{General Embodied Manipulation Benchmarks.}
Prior embodied manipulation benchmarks have progressively advanced standardized evaluation of language-conditioned manipulation, long-horizon control, and generalization across diverse tasks and environments~\cite{james2020rlbench,liu_libero_2023,mees_calvin_2022,li2024evaluating,gu2023maniskill2,robocasa2024,tao2024maniskill3,chen2025robotwin,yu_meta-world_2021}. VIMA includes history-dependent rearrangement and visual constraint satisfaction~\cite{jiang2023vima}, while GENMANIP uses task-oriented scene graphs for constrained scene construction~\cite{gao_genmanip_2025}. These benchmarks already cover several spatial requirements considered here; aggregate task success alone, however, does not distinguish spatial goal grounding errors from execution failures. More specialized benchmarks investigate memory-dependent manipulation. RMBench~\cite{chen2026rmbench} evaluates tasks with varying memory complexity, while RoboMME~\cite{dai2026robomme} examines temporal, spatial, object, and procedural memory. MEMOBench~\cite{sun2026memobench} further introduces process-level evaluations of memory storage, updating, and compression. Together, these benchmarks examine both memory demands and the processes underlying memory-dependent manipulation. \geomind complements them by systematically assessing spatial reasoning and spatial memory through capability-specific manipulation tasks defined by explicit spatial constraints and controlled scene changes. Table~\ref{table:comparison_2} compares selected task requirements and evaluation designs. A checkmark denotes an explicitly targeted task or evaluation subset with reported results; a cross denotes that no explicit evaluation was identified in the reviewed protocol, not that the capability is absent.}

\begin{table}[t]
\centering
\caption{Comparison of spatial capabilities and benchmark design.
\cmark: explicitly evaluated; \xmark: no explicit evaluation identified in the reviewed protocol.}
\label{table:comparison_2}
\begingroup
\scriptsize
\setlength{\tabcolsep}{1.2pt}
\renewcommand{\arraystretch}{0.82}
\newcommand{\auditYes}{\cmark}
\newcommand{\auditNo}{\xmark}
\newcommand{\gheadtight}[1]{%
  {\fontsize{6.2}{6.4}\selectfont\bfseries #1}%
}
\newcommand{\theadittight}[2]{%
  \begin{tabular}[c]{@{}c@{}}
    {\fontsize{5.4}{5.6}\selectfont\itshape #1}\\[-1.2pt]
    {\fontsize{5.4}{5.6}\selectfont\itshape #2}
  \end{tabular}%
}
\begin{tabular*}{\linewidth}{@{\extracolsep{\fill}}lcccccccc@{}}
\toprule
\multirow{2}{*}{\gheadtight{Benchmark}} &
\gheadtight{Design} &
\multicolumn{2}{c}{\gheadtight{Spatial Memory}} &
\multicolumn{4}{c}{\gheadtight{Spatial Reasoning}} &
\gheadtight{Evaluation} \\
\cmidrule(lr){2-2}\cmidrule(lr){3-4}\cmidrule(lr){5-8}\cmidrule(lr){9-9}
& \theadittight{Structure}{Generation}
& \theadittight{Scene}{Updating}
& \theadittight{Identity}{Tracking}
& \theadittight{Target}{Identification}
& \theadittight{Relational}{Grounding}
& \theadittight{Constraint}{Composition}
& \theadittight{Geometric}{Inference}
& \theadittight{Diagnostic}{Metrics} \\
\midrule
VIMA~\cite{jiang2023vima}
& \auditYes & \auditYes & \auditYes & \auditYes & \auditYes & \auditYes & \auditNo & \auditNo \\
GENMANIP~\cite{gao_genmanip_2025}
& \auditYes & \auditNo & \auditNo & \auditNo & \auditYes & \auditNo & \auditNo & \auditNo \\
RoboMME~\cite{dai2026robomme}
& \auditYes & \auditYes & \auditYes & \auditNo & \auditNo & \auditYes & \auditNo & \auditNo \\
MEMOBench~\cite{sun2026memobench}
& \auditYes & \auditYes & \auditNo & \auditNo & \auditYes & \auditNo & \auditNo & \auditYes \\
RoboSPA~\cite{fan2026robospa}
& \auditYes & \auditNo & \auditYes & \auditYes & \auditYes & \auditYes & \auditNo & \auditNo \\
BEHAVIOR-Spatial~\cite{chen2026behaviorspatial}
& \auditNo & \auditNo & \auditNo & \auditYes & \auditYes & \auditYes & \auditNo & \auditYes \\
VABench~\cite{zhang2026vabench}
& \auditYes & \auditYes & \auditNo & \auditNo & \auditYes & \auditNo & \auditNo & \auditYes \\
\midrule
\textbf{\geomind}
& \auditYes & \auditYes & \auditYes & \auditYes & \auditYes & \auditYes & \auditYes & \auditYes \\
\bottomrule
\end{tabular*}

\endgroup
\end{table}

\sure{\textbf{Benchmarks for Spatial Understanding.}
Spatial understanding benchmarks span static image-based evaluations and interactive embodied settings~\cite{zhou2025roborefer,zhao2026espire,yang2024think,mayer2025ivispar,qi_sofar_2025,wen2025dynamicverse}. VSR~\cite{liu2023vsr} evaluates spatial relations in image--text pairs, while other benchmarks examine orientation, navigation, mental rotation, localization, and video recall~\cite{huang2026thinking,stogiannidis2025mind,li2025viewspatial,yu2025visual,STI-Bench,OmniSpatial}. ESI-Bench~\cite{hong2026esibench} emphasizes active information acquisition for spatial reasoning, while SpaMEM~\cite{liao2026spamem} probes spatial state updating and long-horizon belief maintenance through action-conditioned observation sequences. RoboSPA~\cite{fan2026robospa} evaluates spatial reference resolution, compositional spatial selection, and procedural-memory tasks, and reports object-normalized success and progress measures. BEHAVIOR-Spatial~\cite{chen2026behaviorspatial} evaluates spatial instruction following through question answering, target grounding, and robot execution, while VABench~\cite{zhang2026vabench} combines visual demonstrations, active perception, metric control, and trajectory-level diagnostics. \geomind complements these benchmarks by organizing explicit geometric goal construction, current-state re-grounding, and historical identity binding within a shared, controllable tabletop protocol. Structure-preserving sampling, deterministic spatial targets, controlled scene changes, and matched task variants support targeted comparisons across these requirements; scene updating need not imply dependence on unobserved history.}

\sure{\textbf{Spatial Capabilities of Embodied Policies.}
Current embodied agents increasingly rely on vision-language models for language-guided manipulation~\cite{pi0_2024,openvla,bjorck2025gr00t,team2025gemini,gu2025igen}, yet complex spatial instructions remain challenging. Recent approaches therefore incorporate more explicit spatial representations and grounding mechanisms. RoboSpatial~\cite{song2025robospatial} provides large-scale spatial annotations for training vision-language models, while InternVLA-M1~\cite{internvlam1} combines spatial grounding pretraining with spatially guided action learning to connect \textit{where to act} with \textit{how to act}. Other approaches introduce 3D spatial representations to improve the alignment of visual, linguistic, and geometric information for action generation~\cite{qu_spatialvla_2025,zhang_inspire_2025,zhou2025roborefer,li_spatial_2025}. While these methods aim to improve spatial capabilities, \geomind provides a benchmark for systematically examining their strengths and limitations.}

\section{\geomindcolor}
We present the overall design of \geomind by turning structured spatial understanding into a controllable benchmark construction problem. Sec.~\ref{sec:taxonomy} introduces the \geomind taxonomy, and Sec.~\ref{sec:pipeline} describes spatial structure modeling and the benchmark construction pipeline. Sec.~\ref{sec:tasksuite} presents the task suites, showing how executable benchmark episodes are organized into \geomindbase and \geomindaug.
\subsection{Taxonomy of Embodied Spatial Understanding Tasks}
\label{sec:taxonomy}

Rather than categorizing tasks solely by manipulation actions, we organize them according to the spatial capabilities required to determine task-relevant targets and goals. Following the dimensions introduced above, the taxonomy organizes structured spatial understanding around \textit{spatial reasoning} and \textit{spatial memory}.
The former concerns how agents infer targets or goals from spatial relations and scene structure, while the latter concerns how agents maintain and update task-relevant spatial information as scenes change.

\begin{itemize}
    \item \textbf{Target Identification} focuses on static object selection from structured layouts. 
    The agent resolves a spatial description to the correct target object in a fixed scene. 

    \item \textbf{Relational Grounding} requires the agent to realize an explicit spatial relation through action, such as placing one object beside, in front of, or inside another reference object. 

    \item \textbf{Constraint Composition} tests whether the agent can jointly satisfy multiple spatial constraints that define a target object or a valid goal region.

    \item \textbf{Geometric Inference} requires the agent to infer implicit target locations or regions from scene geometry, such as alignment, symmetry, or pattern continuation across objects.

    \item \textbf{Scene Updating} evaluates whether the agent can update task-relevant spatial relations and re-ground its target after objects have moved or been manipulated.

    \item \textbf{Identity Tracking} requires the agent to maintain object identity over time by linking a previously identified object to its current state across scene changes, such as object rearrangement or position exchanges.
\end{itemize}
Fig.~\ref{fig:cap_tasks} qualitatively illustrates the spatial reasoning and memory demands of the six task families and presents representative examples with initial scenes, robot trajectories, and ground-truth spatial targets.
The high/low positions in Fig.~\ref{fig:cap_tasks} indicate intended capability demands.
These tasks remain close to common robotic behaviors, such as selecting, placing, and retrieving objects, while exposing the spatial reasoning and memory demands behind seemingly simple actions.

\subsection{Spatial Structure Modeling and Benchmark Construction}
\label{sec:structure}
\label{sec:pipeline}

\begin{wrapfigure}[17]{r}{0.55\linewidth}
\centering
\includegraphics[width=\linewidth]{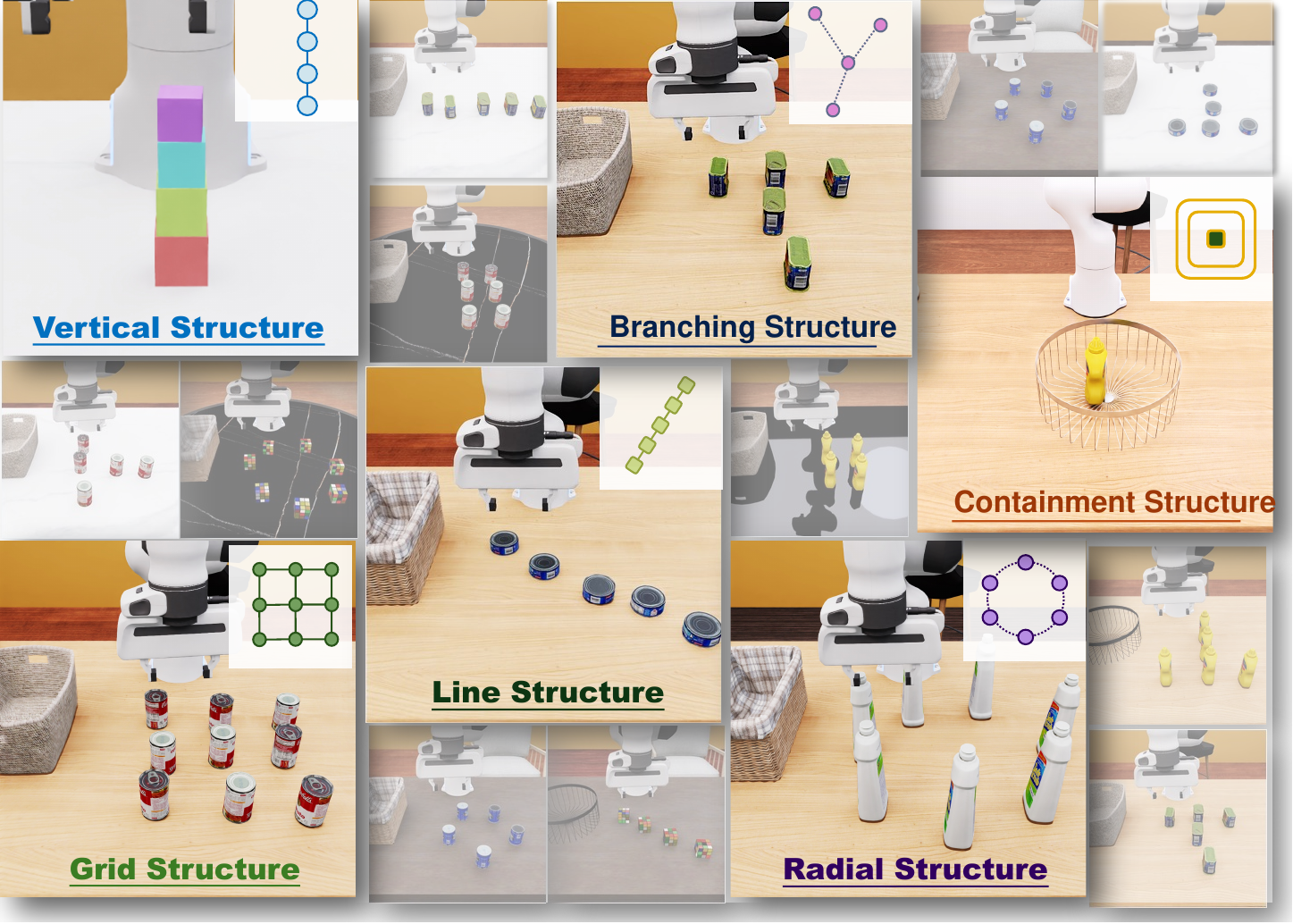}
\vspace{-1.5em}
\caption{Visualization of the six fundamental structural layouts and their derived variants.}
\label{fig:structure}
\vspace{-1.3em}
\end{wrapfigure}

\begin{figure}[t]
\vspace{-1em}
\centering
\includegraphics[width=\linewidth]{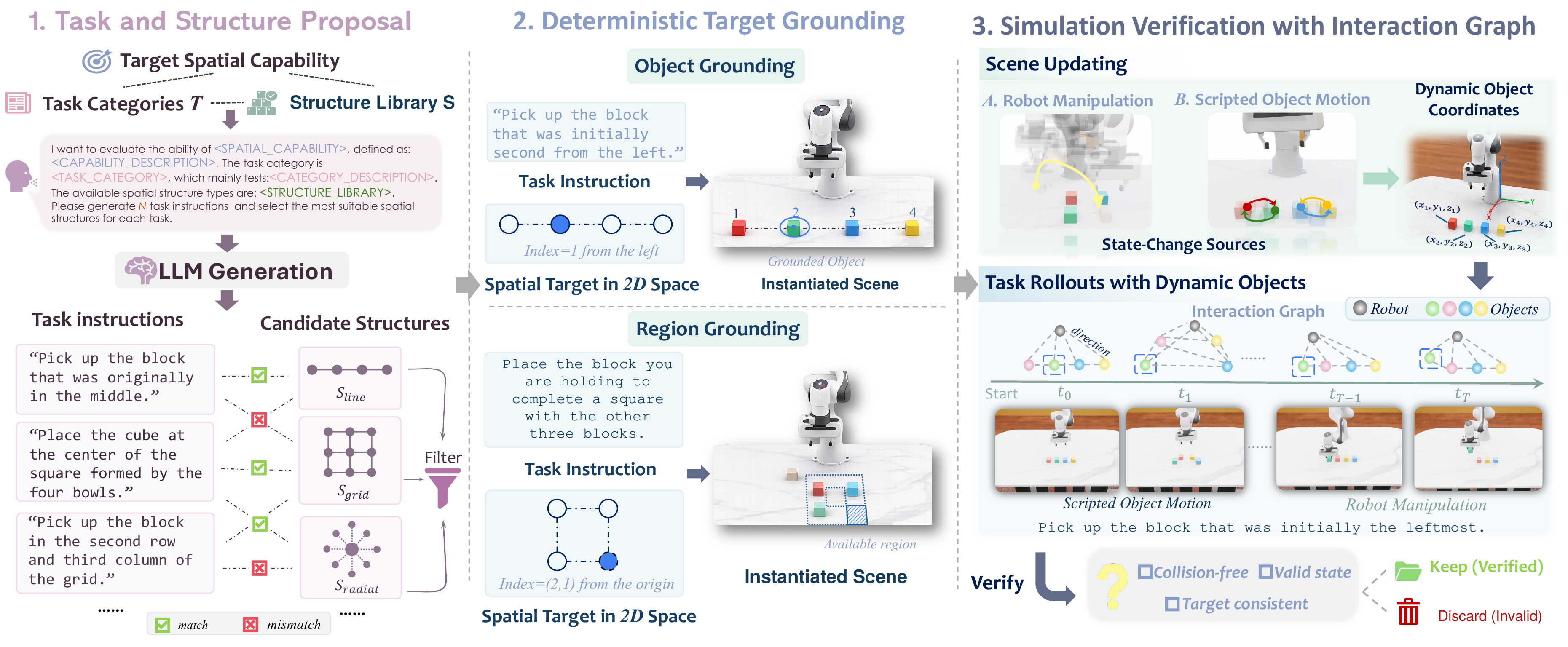}
\caption{\textbf{Overview of the benchmark construction pipeline.} The pipeline first proposes task descriptions and structure templates, then resolves spatial queries over instantiated scenes to derive ground-truth annotations, and finally validates candidate episodes through interaction-graph-based simulation verification.}
\label{fig:pipeline}
\vspace{-1.8em}
\end{figure}

\geomind constructs benchmark episodes through spatial structure modeling and an agent-assisted, rule-verified pipeline. Spatial structures allow object configurations to vary while preserving task-relevant constraints. The pipeline combines task template generation, executable spatial grounding, and simulation-based verification to produce structured manipulation tasks with deterministic ground-truth targets (Fig.~\ref{fig:pipeline}).

\textbf{Spatial Structure Modeling.}
To evaluate spatial understanding across different scene configurations, \geomind represents object layouts using spatial structures that are not tied to a fixed set of coordinates. For example, objects arranged in a line can preserve their left-to-right order while varying in position and spacing.

A task such as selecting the second object from the left therefore follows the same spatial rule across different layouts. We formalize a spatial structure $S$ as a set of relational, topological, and geometric constraints over objects. Given $S$, \geomind generates diverse scene configurations by sampling object poses $\mathbf{X}$ from a structure-preserving distribution:
\[
\mathbf{X} \sim \mathcal{P}(\mathbf{X} \mid S, \epsilon),
\]
where $\epsilon$ controls allowable geometric variations, such as positional jitter and variable spacing, without violating the constraints encoded in $S$. The structure library $\mathcal{S}=\{S_{\mathrm{line}},S_{\mathrm{grid}},\ldots\}$ includes linear, grid, and other canonical layouts (Fig.~\ref{fig:structure}). Sampled configurations must satisfy structural validity and collision-free workspace constraints. This representation supports controlled evaluation across variations in object positions and spacing. Formal definitions are provided in \ref{app:spatial_structure_formalization}.

\textbf{LLM-Assisted Task and Structure Proposal.}
Given a task category and a candidate set of spatial structure types, the LLM proposes matching instructions and structure templates within a predefined schema. Each template specifies the structure type, object roles, layout constraints, sampling ranges, and optional scene-update rules. Candidate templates are screened to remove ambiguous, logically conflicting, or task-incompatible specifications, with human review during template selection. The retained templates form a fixed library for subsequent programmatic scene instantiation.
The benchmark uses a fixed set of tasks selected through human review, with task instances generated programmatically from the corresponding templates.

\textbf{Spatial Target Grounding.}
An LLM-based coding agent translates spatial rules into programs that deterministically resolve target objects or goal regions from simulator states. Directional predicates use the canonical front-camera frame, with object coordinates transformed from the robot base frame. Reasoning tasks ground targets in the current configuration. In memory tasks, instructions precede scripted object motion, and evaluated actions begin after it ends: Scene Updating re-grounds targets in the updated scene, while Identity Tracking preserves their initial identities. Each agent can store earlier observations or maintain an internal memory to identify the target after the objects move.

\textbf{Closed-Loop Verification.}
Each candidate episode is executed in a physics simulator to check physical feasibility and target consistency. For tasks involving scene changes, objects move through scripted motion or simulated robot manipulation. Throughout execution, we record simulator object states and construct a dynamic interaction graph $G_t=(V,E_t)$, where nodes represent objects and edges encode discretized pairwise directions and distances. The graph sequence is used to check whether scene transitions and resulting spatial states are consistent with the task specifications and ground-truth targets. Episodes failing state-validity or target-consistency checks are discarded.

\subsection{Task Suites and Dataset}
\label{sec:tasksuite}

\geomind is organized into two complementary evaluation suites: \geomindbase and \geomindaug.
Together, they provide \textbf{54 evaluated task variants} for structured spatial understanding.
A task variant is a specific test setting that pairs a spatial objective with a scene structure, governed by specific grounding rules and evaluation conditions.
\geomindbase evaluates the six spatial task families under canonical tabletop settings, while \geomindaug introduces controlled perturbations to assess robustness under additional semantic, visual, and execution challenges.

\begin{figure}[t]
    \centering

    \begin{subfigure}[t]{0.25\linewidth}
        \centering
        \includegraphics[width=\linewidth]{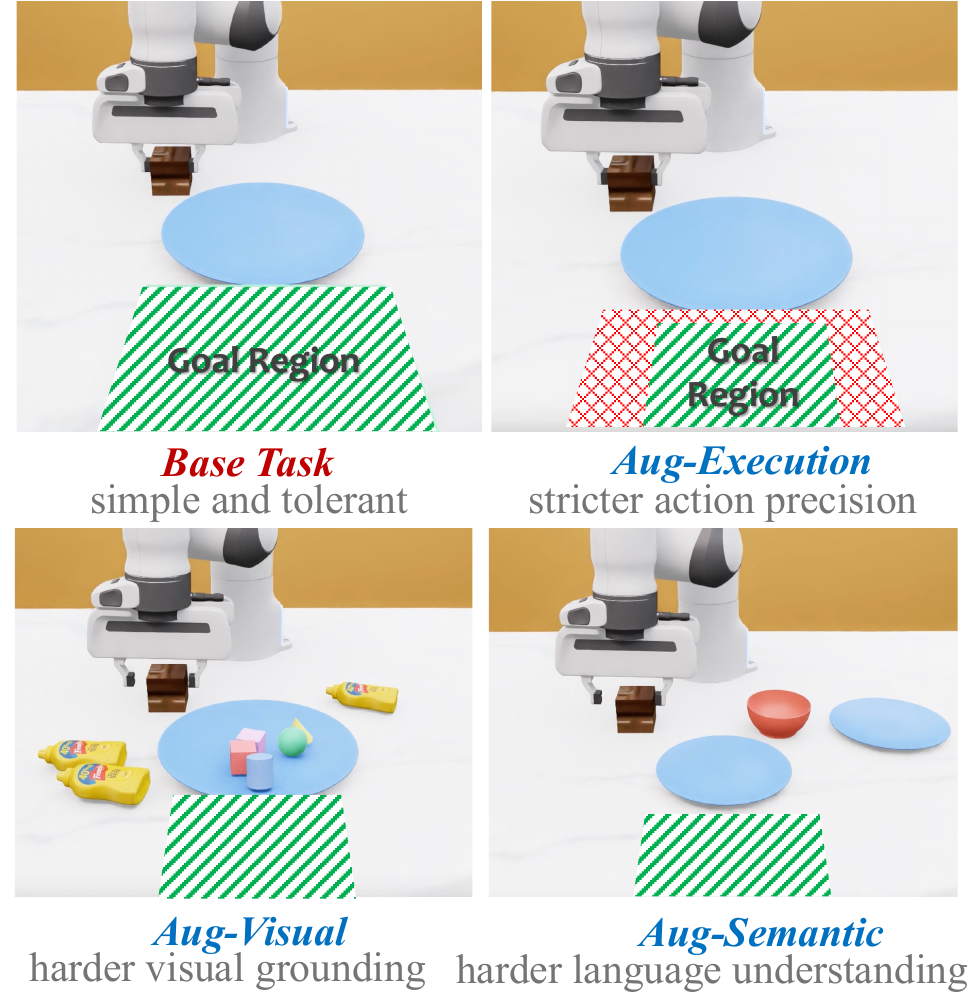}
        \caption{Task display.}
        \label{fig:display}
    \end{subfigure}
    \begin{subfigure}[t]{0.72\linewidth}
        \centering
        \includegraphics[width=\linewidth]{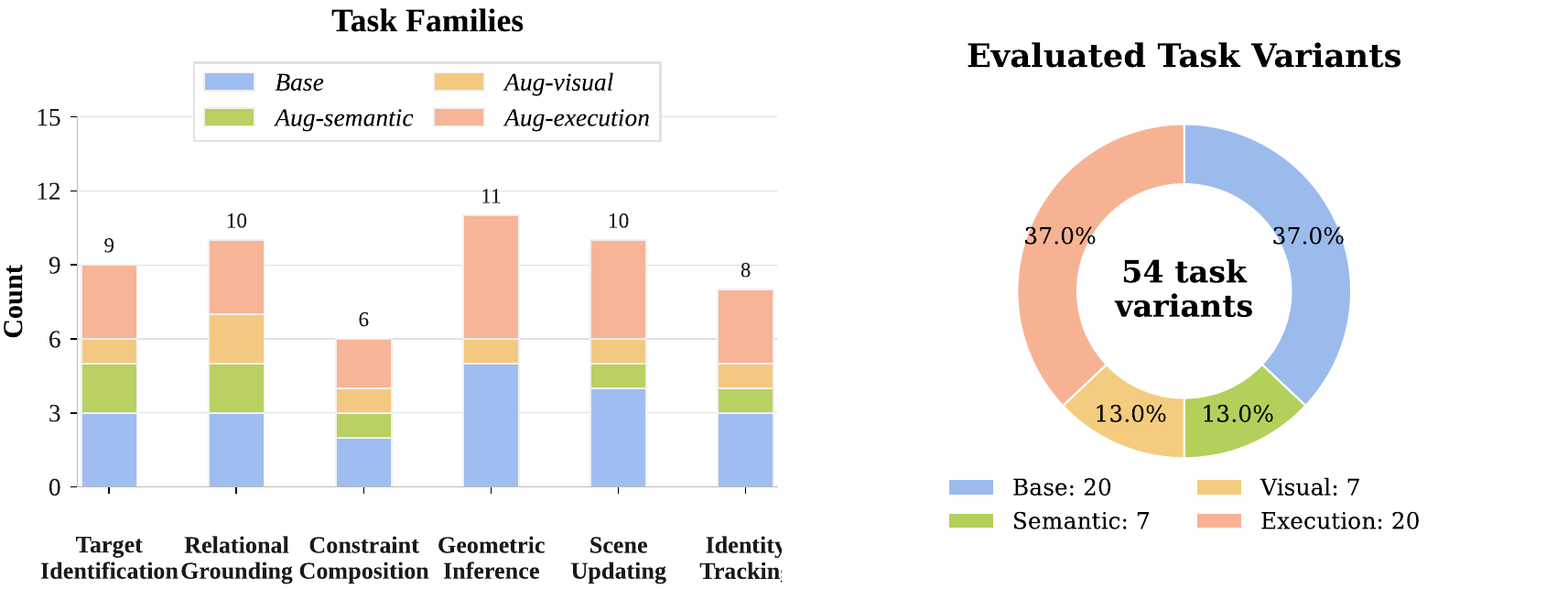}
        \caption{Overview of \geomind task suites. }
        \label{fig:statistics}
    \end{subfigure}

    \caption{Detailed information about \geomind task suites.
    (a) shows how \geomindaug is constructed from \geomindbase.
    (b) shows the distribution of evaluated variants across task families and augmentation types (20 Base, 7 Semantic, 7 Visual, and 20 Execution).    }
    \label{fig:main_ab}

\end{figure}

\geomindbase consists of \textbf{20 evaluated base variants} with diverse language instructions and multiple spatial structure types.
To reduce visual shortcuts, many \geomindbase variants contain multiple visually identical or near-identical objects.
In these settings, target selection or placement depends on spatial cues such as relative position, order, geometric relations, or scene updates, rather than object appearance alone.
\geomindbase serves as the core benchmark suite and evaluates the spatial reasoning and memory capabilities defined in our taxonomy.
Fig.~\ref{fig:statistics} presents the distribution of evaluated task variants across task families and augmentation types.

\geomindaug contains \textbf{34 augmented variants}: 7 Semantic, 7 Visual, and 20 Execution variants. Semantic augmentation modifies language grounding, visual augmentation modifies visual conditions, and execution augmentation tightens action precision requirements. Together with the 20 Base variants, these form 54 evaluated variants. These counts refer to evaluated task variants rather than unique spatial objectives or evaluation episodes. Examples and task details appear in Fig.~\ref{fig:display} and~\ref{app:task_details}.

\section{Experiments}

In this section, we systematically evaluate representative embodied agents on the \geomind benchmark under a controlled setting.
We first describe the experimental setup and the selected embodied agents. Then, we present the evaluation results, analyzing their implications for spatial understanding.

\subsection{Experimental Setup}

\textbf{Benchmarked Embodied Agents.} We evaluate representative embodied agents across \geomindbase and \geomindaug, covering three paradigms: general policies, spatially enhanced policies, and modular agents. These methods differ in whether spatial grounding, planning, and execution are handled implicitly by an end-to-end policy or explicitly by separate modules. Specifically, we evaluate OpenVLA~\cite{openvla}, \(\pi_0\)~\cite{pi0_2024}, and \(\pi_{0.5}\)~\cite{black2025pi} as general policies; SpatialVLA~\cite{qu_spatialvla_2025}, InternVLA-M1~\cite{internvlam1}, and Spatial-Forcing~\cite{li_spatial_2025} as spatially enhanced policies; and COPA~\cite{huang2024copa} and MOKA~\cite{liu2024moka} as modular agents. We additionally include an oracle baseline that uses simulator states and deterministic spatial grounding to compute ground-truth targets, which guide a scripted controller. The oracle provides a reference for assessing task feasibility under correct target information.

\definecolor{avgSecondFill}{HTML}{F6FCF6}
\definecolor{taskSecondFill}{HTML}{FAF6FF}
\renewcommand{\avgSecond}[1]{\cellcolor{avgSecondFill}#1}
\renewcommand{\taskSecond}[1]{\cellcolor{taskSecondFill}#1}

\begin{table}[!t]
\centering
\caption{
Success rates (\%) of different paradigms of embodied agents on \geomindbase.
Excluding the oracle, best and second-best overall results are highlighted with \avgBest{green} shades, while best and second-best family-level results are highlighted with \taskBest{purple} shades.
}
\renewcommand{\arraystretch}{1.15}

\begingroup
\fontsize{10}{12}\selectfont
\setlength{\tabcolsep}{2.8pt}
\resizebox{\textwidth}{!}{%
\begin{tabular}{>{\centering\arraybackslash}p{1.55cm} c c S S S S S S S}
\toprule
\multicolumn{3}{c}{\multirow{2}{*}{\raisebox{-15pt}{\makecell[cc]{\textbf{Embodied Agents}}}}}
& \multicolumn{4}{c}{\textbf{\textit{Spatial Reasoning}}}
& \multicolumn{2}{c}{\textbf{\textit{Spatial Memory}}}
& \multicolumn{1}{c}{\multirow{2}{*}{\raisebox{-15pt}{\makecell[cc]{\textbf{AVG}}}}} \\
\cmidrule(lr){4-7}
\cmidrule(lr){8-9} 
\multicolumn{3}{c}{}
& \multicolumn{1}{c}{\makecell{Target\\Iden.}}
& \multicolumn{1}{c}{\makecell{Relational\\Grounding}}
& \multicolumn{1}{c}{\makecell{Constraint\\Comp.}}
& \multicolumn{1}{c}{\makecell{Geometric\\Inference}}
& \multicolumn{1}{c}{\makecell{Scene\\Updating}}
& \multicolumn{1}{c}{\makecell{Identity\\Tracking}}
& \multicolumn{1}{c}{} \\
\midrule

& \multicolumn{2}{l}{\textbf{Oracle}}
& 100.0 & 98.0 & 100.0 & 100.0 & 98.0 & 100.0 & 99.3 \\

\midrule
\multirow{3}{*}{\makecell[c]{\textbf{General}\\\textbf{Policies}}}
& \multicolumn{2}{c}{OpenVLA}
& \taskSecond{37.5}
& 0.0
& 6.2
& 5.0
& \taskSecond{41.7}
& \taskBest{37.5}
& 20.7 \\

& \multicolumn{2}{c}{$\pi_0$}
& 4.2
& 0.0
& 0.0
& \taskSecond{25.0}
& 6.3
& 20.9
& 11.3 \\

& \multicolumn{2}{c}{$\pi_{0.5}$}
& 4.2 & 0.0 & 0.0 & \taskBest{27.5} & 6.3 & \taskSecond{25.0} & 12.5 \\

\midrule
\multirow{3}{*}{\makecell[c]{\textbf{Spatially}\\\textbf{Enhanced}\\\textbf{Policies}}}
& \multicolumn{2}{c}{SpatialVLA}
& 12.5
& 0.0
& 0.0
& 17.5
& \taskBest{45.8}
& 8.3
& 18.3 \\

& \multicolumn{2}{c}{InternVLA-M1}
& \taskBest{42.5}
& \taskBest{50.0}
& 6.2
& 12.0
& 16.7
& 20.8
& \avgBest{26.4} \\

& \multicolumn{2}{c}{Spatial-Forcing}
& 12.5
& 0.0
& \taskBest{50.0}
& 5.0
& 33.3
& \taskBest{37.5}
& \avgSecond{24.0} \\

\midrule
\multirow{6}{*}{\makecell[c]{\textbf{Modular}\\\textbf{Methods}}}
& \multirow{3}{*}{COPA}
& GPT-4o
& 22.9 & 11.1 & 16.7 & 4.2 & 9.7 & 13.9 & 11.1 \\

&
& Gemini-2.5-Flash
& 10.4 & 2.8 & 6.3 & 1.7 & 6.9 & 0.8 & 4.6 \\

&
& Claude-Sonnet-4.6
& 20.8 & \taskSecond{13.9} & 16.7 & 13.3 & 16.0 & 19.4 & 15.9 \\

\addlinespace[1pt]

& \multirow{3}{*}{MOKA}
& GPT-4o
& 18.8 & 12.5 & \taskSecond{18.8} & 1.7 & 8.3 & 13.9 & 10.1 \\

&
& Gemini-2.5-Flash
& 8.3 & 1.4 & 4.2 & 0.8 & 7.6 & 1.4 & 4.0 \\

&
& Claude-Sonnet-4.6
& 18.8 & \taskSecond{13.9} & 16.7 & 13.3 & 18.6 & 20.8 & 16.9 \\
\bottomrule
\end{tabular}}
\endgroup

\label{tab:main_base_ability}
\end{table}

\textbf{Evaluation Protocol.} All experiments are conducted in the IsaacLab~\cite{isaaclab} simulator using a Franka Research 3 (FR3) robotic arm. All methods are evaluated on the same task instances, random seeds, episode horizons, and success criteria. The evaluated VLA policies use fixed checkpoints without \geomind-specific fine-tuning, and their outputs are adapted to a common action interface. Privileged scene information, including simulator-resolved object identities and ground-truth target coordinates, is used for evaluation and the oracle baseline but is not provided to the evaluated VLA policies or modular agents during inference. Each episode is limited to 1000 control steps and terminates early upon task success or workspace violation. We evaluate \(N_{\mathrm{ep}}=48\) episodes per task variant, with evaluation seeds saved for reproducibility. The physics simulation runs at 480 Hz, and the policy control frequency is 24 Hz. Detailed observation and action interfaces, deployment configurations, and evaluation protocols are provided in~\ref{app:experimental_setup}.

\textbf{Evaluation Metrics.}
We evaluate task completion using Task Success Rate (SR), the proportion of episodes satisfying task-specific success conditions, reported overall, by task family, and by augmentation type. Robustness is measured by $\Delta_{\mathrm{Aug}}=\mathrm{SR}_{\mathrm{Aug}}-\mathrm{SR}_{\mathrm{ref}}$ in percentage points; negative values indicate degradation. The reference is the matched base task for semantic and visual variants, and the corresponding non-execution variant for execution variants.
Spatial Goal Error (SGE) measures final spatial deviation, while Target Tracking Error (TTE) measures average end-effector distance to the target:
\[
\mathrm{SGE}_i=d_i\left(\mathbf{s}_i^{\mathrm{final}},\mathbf{s}_i^{\mathrm{goal}}\right),
\qquad
\mathrm{TTE}_i=\frac{1}{T_i}\sum_{t=1}^{T_i}
\left\|\mathbf{p}_{i,t}^{\mathrm{eef}}-\mathbf{p}_{i,t}^{\mathrm{tar}}\right\|_2.
\]
Here, $i$ indexes episodes, $d_i$ is the evaluator's goal-distance function, $T_i$ is the episode length, and $\mathbf{p}^{\mathrm{eef}}$ and $\mathbf{p}^{\mathrm{tar}}$ denote end-effector and target positions. SGE is aggregated only over valid goal-distance values. TTE uses recorded target trajectories when available, otherwise the initial target position as a marked static-layout proxy. Lower values indicate smaller errors. Applicability and aggregation details are provided in~\ref{app:metric}.

\subsection{Main Results}
We organize the results around five questions concerning overall performance, robustness, failure modes, spatial memory, and spatial reasoning.

\textbf{Q1. How well do embodied agents perform in structured spatial understanding?}

We begin with \geomindbase, which evaluates six spatial capabilities under standard task conditions without the additional perturbations introduced in \geomindaug. Table~\ref{tab:main_base_ability} reports the success rates of different agents across the six task families. The best-performing learned policy, InternVLA-M1, achieves 26.4\% success, followed by Spatial-Forcing at 24.0\%, substantially below the oracle's 99.3\%. Spatially enhanced policies achieve the highest overall scores, although their improvements over general policies are uneven across models and task families. Modular agents achieve 4.0--16.9\% success, indicating that explicit VLM-based grounding and scripted execution also struggle with structured spatial tasks.

To better understand these performance gaps, we examine the contributions of target grounding and execution. The oracle achieves near-perfect success using ground-truth targets and scripted control, demonstrating the feasibility of the benchmark tasks under correct spatial grounding. More directly, in a paired diagnostic evaluation of $\pi_{0.5}$, explicitly specifying the target through its name and a visual cue improves success from 11.3\% to 39.2\%, while keeping the policy, task instances, action interface, and success criterion fixed. This result provides evidence that target identification contributes substantially to the observed failures, although execution and deployment limitations may also play a role.

\textit{\textbf{Finding}: Current embodied agents remain far from reliable on structured spatial manipulation tasks. The controlled target-cue intervention suggests that inaccurate spatial target grounding is an important contributor to these failures, beyond execution limitations alone.}

\textbf{Q2. How robust is structured spatial understanding under semantic, visual, and execution perturbations?}

{\hyphenpenalty=10000
\exhyphenpenalty=10000
\sloppy
Table~\ref{tab:aug_drop_half} shows that success in \geomindbase is fragile under augmented variants of the same spatial goals.
The clearest drop comes from semantic augmentation: averaged over the six embodied policies, success decreases from about 22.6\% on the corresponding Base tasks to about 14.2\%, and every policy degrades.
This indicates that many Base successes still rely on simple instruction grounding rather than robust spatial-language understanding.
The effects of visual and execution augmentation are more uneven.
OpenVLA falls to 12.5\% under both settings, whereas $\pi_0$ remains unchanged under visual perturbation and decreases by 6.5 percentage points under execution.  These results suggest that structured spatial understanding degrades under complex instruction semantics and increased visual noise.
\par}

\textit{\textbf{Finding}: Spatial success in embodied agents is condition-dependent rather than goal-invariant: the same spatial objective often fails once its linguistic, visual, or execution support is perturbed.}

\begin{table}[!htbp]
\centering
\begin{minipage}[t]{0.80\textwidth}
\vspace{0pt}
    \centering
  \setlength{\belowcaptionskip}{4pt}
  \caption{Success rates (\%) on \geomindaug and the reported base references. Parentheses show changes in percentage points.}

  \label{tab:aug_drop_half}
  \begingroup
  \fontsize{9}{12}\selectfont
  \setlength{\tabcolsep}{1pt}
  \renewcommand{\arraystretch}{0.92}

  \providecolor{NatureMutedPink}{HTML}{F7EFF2}
  \colorlet{augcol}{NatureMutedPink}
  \colorlet{negcol}{red!75!black}
  \colorlet{poscol}{black!55}

  \newcommand{\augneg}[2]{#1\,{\footnotesize\textcolor{negcol}{($-#2$)}}}
  \newcommand{\augpos}[2]{#1\,{\footnotesize\textcolor{poscol}{($+#2$)}}}

  \begin{tabular*}{\linewidth}{@{\extracolsep{\fill}}l l c >{\columncolor{augcol}}c >{\columncolor{augcol}}c >{\columncolor{augcol}}c@{}}
    \toprule
    \multicolumn{2}{c}{\textbf{Policies}}
    & \textbf{Corresponding Base}
    & \textbf{Semantic}
    & \textbf{Visual}
    & \textbf{Execution} \\
    \midrule

    \multicolumn{2}{l}{OpenVLA}
    & 25.0
    & \augneg{16.6}{8.4}
    & \augneg{12.5}{12.5}
    & \augneg{12.5}{12.5} \\

    \multicolumn{2}{l}{$\pi_0$}
    & 16.7
    & \augneg{12.5}{4.2}
    & \augpos{16.7}{0.0}
    & \augneg{10.2}{6.5} \\

    \multicolumn{2}{l}{$\pi_{0.5}$}
    & 18.5 & \augneg{12.5}{6.0} &\augneg{16.7}{1.8} &\augneg{10.2}{8.3} \\

    \midrule
    \multicolumn{2}{l}{SpatialVLA}
    & 18.8
    & \augneg{8.3}{10.5}
    & \augneg{12.5}{6.3}
    & \augneg{12.5}{6.3} \\

    \multicolumn{2}{l}{InternVLA-M1}
    & 37.5
    & \augneg{27.1}{10.4}
    & \augpos{37.5}{0.0}
    & \augneg{35.4}{2.1} \\

    \multicolumn{2}{l}{Spatial-Forcing}
    & 18.8
    & \augneg{8.3}{10.5}
    & \augneg{12.5}{6.3}
    & \augneg{16.7}{2.1} \\








    \bottomrule
  \end{tabular*}
  \endgroup
  \end{minipage}
\end{table}

\textbf{Q3. Why do embodied agents fail on \geomind?}

\begin{figure}[!h]
\centering

\begin{subfigure}[t]{0.76\linewidth}
    \centering
    \includegraphics[width=0.98\linewidth]{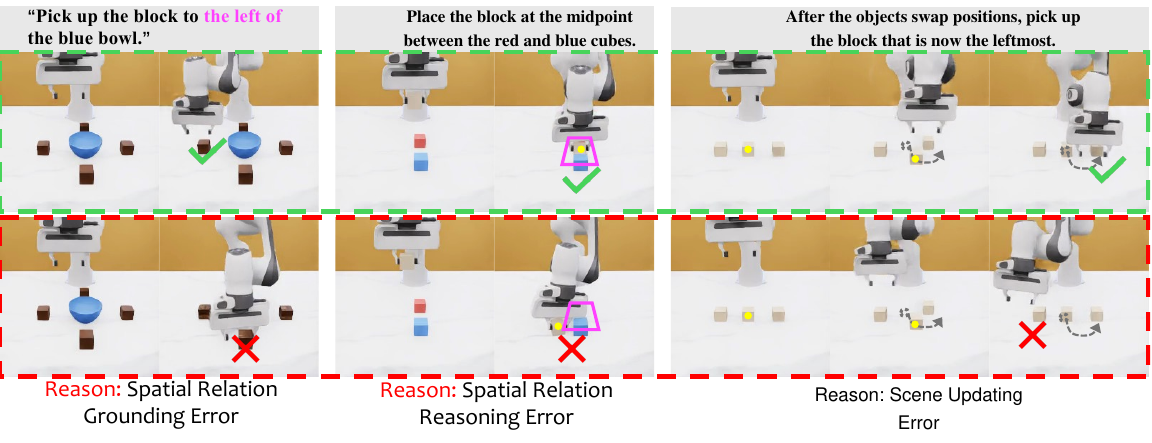}
    \caption{Embodied policies.}
    \label{fig:vla_failure}
\end{subfigure}
\hfill
\begin{subfigure}[t]{0.22\linewidth}
    \centering
    \includegraphics[width=0.98\linewidth]{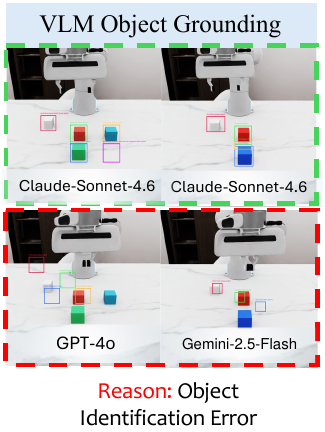}
    \caption{Modular agents.}
    \label{fig:vlm_failure}
\end{subfigure}
\vspace{-0.5em}
\caption{
Visualization of typical failure cases in \geomind.
}
\label{fig:failure}
\end{figure}

We examine representative failure cases in Fig.~\ref{fig:failure} to connect the quantitative trends with concrete trajectories.
For embodied policies in Fig.~\ref{fig:vla_failure}, the illustrated failures fall into three categories: \textit{spatial relation grounding errors}, where policies bind the correct anchor but execute the wrong relation; \textit{spatial relation reasoning errors}, where policies fail to infer structure-defined targets; and \textit{scene-updating errors}, where policies fail to ground the requested target in the updated configuration. These examples suggest that some failures involve converting spatial observations into appropriate action targets, although the trajectories alone do not isolate grounding errors from control errors.
For modular embodied agents in Fig.~\ref{fig:vlm_failure}, the illustrated failures include object localization errors that can propagate to planning and control. This suggests a different bottleneck: modular agents may expose explicit reasoning steps, but remain vulnerable when intermediate spatial grounding is inaccurate.

\textit{\textbf{Finding}: The illustrated failures suggest difficulties in translating spatial observations into correct action targets. These examples identify possible failure mechanisms, but do not establish their prevalence across the benchmark.}

\textbf{Q4. What makes spatial memory difficult for embodied agents?}

\begin{table}[!htbp]
    \centering
  \fontsize{9}{12}\selectfont
  \setlength{\belowcaptionskip}{4pt}
  \caption{Family-level failure rates and auxiliary spatial errors. Lower values are better.}
  \label{tab:ability_diagnostics_with_tte}
  \setlength{\tabcolsep}{2.2pt}
  \renewcommand{\arraystretch}{0.92}
  \begin{tabular*}{0.90\linewidth}{@{\extracolsep{\fill}}llccc@{}}
    \toprule
    \textbf{Capability Dimensions}
    & \textbf{Task Family}
    & \textbf{Failure Rate \scalebox{0.9}{$\downarrow$}}
    & \textbf{SGE (m) \scalebox{0.9}{$\downarrow$}}
    & \textbf{TTE (m) \scalebox{0.9}{$\downarrow$}} \\
    \midrule
    \multirow{2}{*}{Spatial Memory}
    & Scene Updating
    & 68.89\%
    & 0.189
    & 0.330 \\
    & Identity Tracking
    & 85.00\%
    & 0.192
    & 0.486 \\
    \midrule
    \multirow{4}{*}{Spatial Reasoning}
    & Target Identification
    & 72.08\%
    & 0.116
    & 0.811 \\
    & Relational Grounding
    & 98.33\%
    & 0.180
    & 0.672 \\
    & Constraint Composition
    & 87.50\%
    & 0.142
    & 0.627 \\
    & Geometric Inference
    & 86.67\%
    & 0.158
    & 0.727 \\
    \bottomrule
  \end{tabular*}
  \end{table}

Tables~\ref{tab:main_base_ability} and~\ref{tab:ability_diagnostics_with_tte} show that agents struggle with tasks involving scene changes. Scene Updating requires selecting a target from the updated configuration, whereas Identity Tracking requires linking an object identified before the change to its new position. An agent may therefore fail an identity-tracking task by selecting an object that matches the current spatial relation but is not the original target. This provides a possible explanation for failures beyond imprecise execution. However, aggregate SGE and TTE cannot distinguish these causes: both are affected by scene geometry, initial target distance, and rollout duration. We use them as auxiliary diagnostics rather than direct measures of identity preservation or evidence that one task family is intrinsically harder.

\textit{\textbf{Finding}: Identity Tracking requires preserving object correspondence across scene changes; the current aggregate results do not establish how much failure is attributable to incorrect target selection versus execution.}

\textbf{Q5. What makes spatial reasoning difficult for embodied agents?}

Table~\ref{tab:ability_diagnostics_with_tte} reports the highest failure rate for relational grounding among the reasoning families in that diagnostic summary, at 98.33\%, compared with 72.08\% for target identification. Constraint composition and geometric inference are also difficult, with failure rates of 87.50\% and 86.67\%, respectively.
This indicates that current agents do not only struggle with recognizing spatial targets; they also fail to convert explicit spatial relations into correct embodied actions.
Table~\ref{tab:main_base_ability} shows that geometric inference has the lowest best-agent success rate, with the strongest method reaching only 27.5\%, suggesting that implicit structure-based reasoning remains hard even for the best agents.
Together, these results suggest that agents remain weak when they must jointly satisfy multiple spatial constraints or infer targets from implicit geometric structure.

\textit{\textbf{Finding}: Spatial reasoning is difficult not only because agents struggle to identify the correct spatial target, but because they fail to transform spatial constraints into executable manipulation behavior.}

\section{Conclusion}
We present \geomind, a systematic benchmark for evaluating spatial understanding in embodied agents. By organizing tasks around spatial memory and spatial reasoning, \geomind enables a more fine-grained analysis than overall task success alone. We further emphasize how spatial structure underpins spatial understanding, and present an automated generation pipeline that allows scalable benchmark construction.
Our experiments show that current embodied agents remain far from achieving consistently high success rates on \geomind, with large performance gaps and highly uneven abilities across task families.  These findings suggest that spatial understanding remains a core limitation of embodied agents.

\paragraph{Limitations.}
\geomind is limited to simulated, single-arm tabletop manipulation, without real-robot validation. Its controlled layouts do not capture the full diversity of real-world environments. Heavy occlusion, viewpoint changes, sensor noise, and complex contact dynamics are not systematically evaluated, although incidental occlusion may occur during execution. Semantic and visual augmentations cover selected transformations rather than exhaustive robustness conditions. Our experiments evaluate frozen pretrained policies, and their performance may also depend on observation and action-interface compatibility. We do not report benchmark-specific fine-tuning results; learning spatial skills and generalizing to held-out structures or variants remain untested. The results therefore characterize frozen-policy performance in the evaluated setting and do not establish transfer to real robots.

\section*{Code and Data Availability}
The benchmark code and data have not yet been publicly released. We plan to release the benchmark assets, evaluation code, configuration files, and reproduction instructions in the project repository:
\url{https://github.com/Notfound-JH/GEOMIND-benchmark.git}.

\begin{ack}
\textbf{Funding Statement.} This research received no specific grant from any funding agency in the public, commercial, or not-for-profit sectors.

\textbf{Competing Interests.} The authors declare no competing interests.
\end{ack}

\bibliographystyle{unsrt}
\bibliography{neurips_2026}
\newpage
\appendix
\renewcommand{\thesection}{Appendix \Alph{section}}
\renewcommand{\thesubsection}{\Alph{section}.\arabic{subsection}}
\section{Metrics Formulation}
\label{app:metric}
\subsection{Success Rate}
\label{app:suc_rate}
We compute the overall and family success rates as
\[
\mathrm{SR}_{\mathrm{All}}
=
\frac{1}{|\mathcal{T}|N_{\mathrm{ep}}}
\sum_{\tau\in\mathcal{T}}
\sum_{i=1}^{N_{\mathrm{ep}}}
\mathbb{I}[s_{\tau,i}=1],
\quad
\mathrm{SR}_{f}
=
\frac{1}{|\mathcal{T}_{f}|N_{\mathrm{ep}}}
\sum_{\tau\in\mathcal{T}_{f}}
\sum_{i=1}^{N_{\mathrm{ep}}}
\mathbb{I}[s_{\tau,i}=1],
\]
where $s_{\tau,i}$ indicates whether episode $i$ of task variant $\tau$ succeeds, $\mathcal{T}$ denotes the set of evaluated task variants, and $\mathcal{T}_{f}$ denotes the subset in family $f$. Each variant has $N_{\mathrm{ep}}=48$ episodes. With equal episode counts per variant, this aggregation weights task variants equally rather than weighting task families equally.

\subsection{Spatial Goal Error}
\label{app:sge}

Spatial Goal Error (SGE) is an auxiliary diagnostic metric that measures the final distance between the achieved spatial state and the target spatial goal. It is designed to capture final spatial precision, rather than full task completion.

For the $i$-th episode, we define
\begin{equation}
\mathrm{SGE}_i =
d_i\left(
\mathbf{s}^{\mathrm{final}}_i,
\mathbf{s}^{\mathrm{goal}}_i
\right),
\end{equation}
where $\mathbf{s}^{\mathrm{final}}_i$ denotes the final achieved state and $\mathbf{s}^{\mathrm{goal}}_i$ denotes the target spatial state, and $d_i$ is the evaluator's goal-distance function for episode $i$. For tasks with an explicit target position, this reduces to the Euclidean distance
\begin{equation}
\mathrm{SGE}_i =
\left\|
\mathbf{p}^{\mathrm{final}}_i -
\mathbf{p}^{\mathrm{goal}}_i
\right\|_2 .
\end{equation}

In our implementation, SGE is obtained from the evaluator-provided goal-distance proxy, such as \texttt{evaluation.goal\_distance} or \texttt{distance\_m}. If an episode does not provide a valid goal-distance value, its SGE is treated as missing and excluded from SGE aggregation. Therefore, whenever SGE is reported, we also record the number of valid SGE episodes.

For a model $m$ and task family $f$, we aggregate SGE over valid episodes:
\begin{equation}
\overline{\mathrm{SGE}}_{m,f}
=
\frac{1}{|\mathcal{I}^{\mathrm{SGE}}_{m,f}|}
\sum_{i \in \mathcal{I}^{\mathrm{SGE}}_{m,f}}
\mathrm{SGE}_i ,
\end{equation}
where $\mathcal{I}^{\mathrm{SGE}}_{m,f}$ denotes the set of episodes with valid SGE values for model $m$ and family $f$.

For a specified set $\mathcal{M}$ of evaluated models, we first compute the model-family average and then average over models:
\begin{equation}
\overline{\mathrm{SGE}}_{f}
=
\frac{1}{|\mathcal{M}|}
\sum_{m \in \mathcal{M}}
\overline{\mathrm{SGE}}_{m,f}.
\end{equation}
These averages are defined only for nonempty valid-episode sets. A model-family pair with no valid values has an undefined SGE and must not be treated as zero; the cross-model formula applies when each model in the specified set has a defined value. This weighting avoids giving models with more valid episodes more weight.

\noindent\textbf{Interpretation.}
Lower SGE indicates better final spatial precision. However, SGE does not measure whether the agent followed the correct trajectory, selected the correct object throughout the rollout, or satisfied all spatial relations. It is therefore used only as an auxiliary diagnostic, not as a replacement for official task success.
\subsection{Target Tracking Error}
\label{app:tte}

Target Tracking Error (TTE) is an auxiliary trajectory-level diagnostic metric. While SGE measures final spatial error, TTE measures whether the end-effector remains close to the intended target throughout the rollout. It measures target proximity along the trajectory, but does not directly measure spatial memory or identity preservation.

For the $i$-th episode with trajectory length $T_i$, we define
\begin{equation}
\mathrm{TTE}_i =
\frac{1}{T_i}
\sum_{t=1}^{T_i}
\left\|
\mathbf{p}^{\mathrm{eef}}_{i,t}
-
\mathbf{p}^{\mathrm{tar}}_{i,t}
\right\|_2 ,
\end{equation}
where $\mathbf{p}^{\mathrm{eef}}_{i,t}$ is the end-effector position at timestep $t$, and $\mathbf{p}^{\mathrm{tar}}_{i,t}$ is the ground-truth target position at the same timestep.

When dynamic object poses are available in the trajectory trace, the target position is taken from the target object's dynamic pose:
\begin{equation}
\mathbf{p}^{\mathrm{tar}}_{i,t}
=
\mathbf{p}^{\mathrm{obj}}_{i,t}.
\end{equation}
When dynamic object poses are unavailable, we use the static target position from the sampled layout as a proxy:
\begin{equation}
\mathbf{p}^{\mathrm{tar}}_{i,t}
=
\mathbf{p}^{\mathrm{tar}}_{i,0}.
\end{equation}
In this case, the metric is marked as a static-layout proxy.

For a model $m$ and task family $f$, TTE is first computed at the episode level and then averaged over valid episodes:
\begin{equation}
\overline{\mathrm{TTE}}_{m,f}
=
\frac{1}{|\mathcal{I}^{\mathrm{TTE}}_{m,f}|}
\sum_{i \in \mathcal{I}^{\mathrm{TTE}}_{m,f}}
\mathrm{TTE}_i ,
\end{equation}
where $\mathcal{I}^{\mathrm{TTE}}_{m,f}$ denotes the set of episodes with valid trajectory traces for model $m$ and family $f$.

When reporting a family-level average across models, we compute
\begin{equation}
\overline{\mathrm{TTE}}_{f}
=
\frac{1}{|\mathcal{M}|}
\sum_{m \in \mathcal{M}}
\overline{\mathrm{TTE}}_{m,f}.
\end{equation}
The same nonempty-set requirement applies to TTE: a model-family pair without a valid trajectory has an undefined value, and the cross-model formula requires a defined value for each model in the specified set. This model-wise aggregation prevents a run with more episodes from dominating the reported value. We avoid pooling all timesteps across episodes directly, because such pooling would give longer trajectories larger weight.

\noindent\textbf{Interpretation.}
Lower TTE means that the end-effector stays closer to the intended target over the trajectory. Unlike SGE, which only evaluates the final state, TTE captures trajectory-level target following. However, TTE still does not directly verify successful grasping, placement, or spatial-relation satisfaction, so it is used as an auxiliary diagnostic rather than an official success metric.

\section{Detailed Experimental Setup}
\label{app:experimental_setup}

This section provides the implementation details needed to reproduce the \geomind evaluation.
We focus on the experimental variables that directly affect validity and reproducibility:
simulator and robot setup, control frequency, policy inputs, action conversion, rollout protocol,
baseline deployment, and information access. Unless otherwise stated, all methods are evaluated
on the same task instances, random seeds, workspace constraints, episode horizon, and success
criteria.

\subsection{Simulation, Robot, and Control}

All experiments are conducted in \textsc{GeoMind Lab}, an IsaacLab-based benchmark package built on
Isaac Sim. We use a single Franka Research 3 (FR3) arm equipped with the FR3 parallel-jaw gripper.
The robot receives relative end-effector commands and is controlled by a differential
inverse-kinematics controller. The gripper is controlled through finger joint-position commands and
is exposed to policies as a binary open/close command.

The physics simulation runs at $480$ Hz, while policies are queried at $24$ Hz. Each policy action is
therefore executed through the low-level controller over 20 physics substeps. An episode
terminates when the task succeeds, the robot violates the workspace guard, or the maximum horizon is
reached. Table~\ref{tab:sim_control_setup} summarizes the main simulator and control settings.

\begin{table}[t]
\centering
\caption{
Simulation and control settings used for all evaluations.
}
\label{tab:sim_control_setup}
\setlength{\tabcolsep}{7pt}
\renewcommand{\arraystretch}{1.12}
\begin{tabular}{ll}
\toprule
\textbf{Item} & \textbf{Setting} \\
\midrule
Simulator & Isaac Sim / IsaacLab \\
Benchmark package & \textsc{GeoMind Lab} \\
Robot & Franka Research 3 (FR3) \\
Gripper & FR3 parallel-jaw gripper \\
Low-level controller & Differential inverse kinematics \\
Physics frequency & $480$ Hz \\
Policy control frequency & $24$ Hz \\
Episode horizon & $1000$ control steps \\
Termination & Success / workspace violation / horizon \\
Workspace guard $(x,y,z)$
& $[0.15,0.75]\times[-0.45,0.45]\times[0.70,1.20]$ m \\
\bottomrule
\end{tabular}
\end{table}

\subsection{Observation and Action Interface}

All end-to-end VLA policies are evaluated through a common observation and action interface. The
policy input contains the language instruction, two RGB camera observations, and robot
proprioception. The two cameras consist of a static table-view camera and a wrist-mounted camera,
both rendered at $224\times224$ resolution.

Importantly, privileged simulator information is not exposed to VLA policies. Object poses,
object identities, segmentation masks, ground-truth task variables, oracle target coordinates, and
task metadata are used only for evaluation. Table~\ref{tab:obs_action_interface} summarizes the policy
interface and the evaluation-only states.

\begin{table}[t]
\centering
\caption{
Observation and action interface. Privileged simulator states are excluded from VLA policy inputs.
}
\label{tab:obs_action_interface}
\setlength{\tabcolsep}{5.5pt}
\renewcommand{\arraystretch}{1.12}
\begin{tabular}{lll}
\toprule
\textbf{Type} & \textbf{Field} & \textbf{Description} \\
\midrule
\multirow{5}{*}{Policy input}
& Instruction & Natural-language task instruction \\
& Table-view RGB & Static external camera observation, $224\times224$ \\
& Wrist RGB & Wrist-mounted camera observation, $224\times224$ \\
& Joint state & Robot joint positions and velocities \\
& End-effector / gripper state & Current end-effector pose and gripper state \\
\midrule
\multirow{3}{*}{Policy action}
& Translation & Relative Cartesian displacement \\
& Rotation & Relative rotation-vector command \\
& Gripper & Binary open/close command \\
\midrule
\multirow{5}{*}{Evaluation only}
& Object poses & Used for success checking and diagnostics \\
& Object identities & Used for ground-truth target resolution \\
& Segmentation masks & Not exposed to VLA policies \\
& Task metadata & Used only for evaluation and diagnostics \\
& Target regions / coordinates & Not exposed to VLA policies \\
\bottomrule
\end{tabular}
\end{table}

The common action interface is a 7-DoF relative end-effector command,
\[
\mathbf{a}_t =
[\Delta x,\Delta y,\Delta z,\Delta r_x,\Delta r_y,\Delta r_z,g],
\]
where the first three dimensions represent Cartesian displacement, the next three dimensions
represent a relative rotation-vector command, and the final dimension controls the gripper. For
policies whose native output format differs from this interface, we convert the output to the same
7-DoF command before execution.

We use a unified normalized action protocol. Each policy output is first converted to a 7-DoF action
vector and clipped to $[-1,1]$. The translational and rotational components are then scaled by a
model-specific multiplier $s_m$. In the following equations, $a$ denotes the converted, already clipped action vector. Slices use zero-based indexing with an exclusive upper bound: $a_{0:3}$ contains the three translation components and $a_{3:6}$ the three rotation components. The $\mathrm{clip}$ operator acts componentwise:

\[
u_{xyz} = \mathrm{clip}(s_m a_{0:3}, -1, 1),
\]
\[
u_{rot} =
\begin{cases}
0, & \text{if rotation is disabled for the model},\\
\mathrm{clip}(2s_m a_{3:6}, -1, 1), & \text{otherwise}.
\end{cases}
\]
The controller converts these normalized components to physical increments using scales of 0.5 m for translation and 0.5 rad for rotation, yielding
\[
\Delta p = 0.5 u_{xyz}\ \mathrm{m},
\qquad
\Delta r = 0.5 u_{rot}\ \mathrm{rad}.
\]
The gripper command is thresholded at zero: positive values are mapped to open and negative values
are mapped to close. For baselines without reliable gripper outputs, we keep the gripper fixed
according to their default deployment protocol and report this setting explicitly.

\subsection{Evaluation Protocol}

We evaluate each method on the same task instances and random seeds. Each evaluated task variant contains $48$
evaluation episodes. \geomindbase contains $20$ variants, resulting in $960$ episodes per method.
\geomindaug contains $34$ augmented variants ($7$ Semantic, $7$ Visual, and $20$ Execution), resulting in $1632$ episodes per method, for $2592$ episodes across both suites. Success rate is
the primary metric and is computed over all episodes in the corresponding task suite or task family.

\begin{table}[t]
\centering
\caption{
Evaluation protocol for \geomindbase and \geomindaug.
}
\label{tab:evaluation_protocol}
\small
\setlength{\tabcolsep}{7pt}
\renewcommand{\arraystretch}{1.12}
\begin{tabular}{ll}
\toprule
\textbf{Item} & \textbf{Setting} \\
\midrule
\geomindbase & $20$ variants, $48$ episodes per variant, $960$ episodes per method \\
\geomindaug & $34$ augmented variants, $48$ episodes per variant, $1632$ episodes per method \\
Task instances & Shared across methods \\
Random seeds & Shared across methods \\
Episode horizon & $1000$ control steps \\
Policy control frequency & $24$ Hz \\
Physics frequency & $480$ Hz \\
Camera resolution & $224\times224$ \\
Primary metric & Success rate \\
VLA adaptation & No \geomind-specific fine-tuning \\
\bottomrule
\end{tabular}
\end{table}

\subsection{VLA Baseline Deployment}
\label{app:vla_deployment}

All VLA baselines are evaluated as frozen policies without \geomind-specific fine-tuning. Each model is
deployed as a remote policy server and queried online during rollout. For each model, we use a fixed
checkpoint and server configuration across all evaluated episodes.

The policy server receives only the allowed policy inputs described in
Table~\ref{tab:obs_action_interface}. No VLA baseline receives privileged simulator object poses,
segmentation masks, ground-truth task variables, oracle target coordinates, or task metadata through
the policy interface.

Some VLA models return action chunks rather than a single action. In our deployment, the client
executes only the first valid action from the returned chunk and then queries the policy again at the
next control step. This keeps all policies closed-loop at the shared $24$ Hz control frequency while
preserving each model's native action-generation interface.

Different VLA models use different native action heads and deployment conventions. We therefore
report the model-specific action multiplier, checkpoint setting, rotation setting, and gripper
setting in Table~\ref{tab:vla_deployment}. These settings are fixed for a given model across the full
evaluation. Simulations were conducted on a workstation equipped with an NVIDIA RTX 3090 GPU, while model inference was performed on an NVIDIA RTX 4090 (48 GB).

\begin{table}[t]
\centering
\caption{
Deployment settings of VLA baselines. $s_m$ denotes the normalized action gain used before controller-level command limiting. All models are evaluated with fixed checkpoints and without
\geomind-specific fine-tuning.
}
\label{tab:vla_deployment}
\setlength{\tabcolsep}{5.5pt}
\renewcommand{\arraystretch}{1.12}
\begin{tabular}{lcccl}
\toprule
\textbf{Model}
& \textbf{Fine-tuning}
& \textbf{Checkpoint}
&  $s_m$
& \textbf{Rotation / gripper} \\
\midrule
OpenVLA         & No & Fixed & $2.0$ & Rotation enabled; binary gripper \\
$\pi_0$           & No & Fixed & $0.5$ & Rotation disabled; fixed gripper \\
$\pi_{0.5}$       & No & Fixed & $0.5$ & Rotation disabled; fixed gripper \\
SpatialVLA       & No & Fixed & $5.0$ & Rotation enabled; binary gripper \\
InternVLA-M1    & No & Fixed & $0.5$ & Rotation enabled; binary gripper \\
Spatial-Forcing  & No & Fixed & $0.2$ & Rotation enabled; binary gripper \\
\bottomrule
\end{tabular}
\end{table}

\subsection{Modular Baselines and Diagnostic Use}
\label{app:modular_agents}

We additionally evaluate modular agents based on COPA~\cite{huang2024copa} and MOKA~\cite{liu2024moka}. 
These methods represent a different embodied-agent paradigm from end-to-end VLA policies: instead of directly predicting low-level actions from visual observations and language instructions, they decompose manipulation into explicit visual-language grounding, target selection, and scripted execution. 
We include them as modular baselines to compare different design paradigms for structured spatial manipulation, and also use their intermediate predictions to diagnose where failures arise.

Each modular agent receives the task instruction and RGB observations, and uses a VLM backbone to identify the task-relevant object or spatial target from the scene. 
We instantiate the modular agents with three VLM backbones: GPT-4o, Gemini-2.5-Flash, and Claude-Sonnet-4.6. 
The predicted target is then converted into Cartesian waypoints or a target region and executed with the same Franka differential-IK controller used in the VLA evaluation. 
For pick-and-place tasks, the scripted routine follows a top-down sequence of approach, grasp, lift, place, release, and retreat. 
For reaching tasks, the controller moves the end effector toward the predicted target region and evaluates success using the same task evaluator as other methods.

All modular agents are evaluated on the same task instances, random seeds, workspace constraints, episode horizon, and success criteria as the end-to-end policies. 
They do not receive privileged simulator information during inference: oracle target coordinates, simulator-resolved object identities, ground-truth task variables, segmentation masks, and task metadata are not exposed to the VLM or the planner. 
These privileged states are used only by the evaluator for success checking and diagnostic logging.

Because modular agents use a different inference and execution structure from VLA policies, their results should be interpreted as a comparison across embodied-agent paradigms rather than as a fully controlled comparison of identical policy interfaces. 
This distinction is important: end-to-end policies test whether a learned visuomotor model can directly map observations and language to actions, whereas modular agents test whether explicit VLM-based grounding and scripted execution can solve the same spatial objectives. 
Their low success rates therefore provide complementary evidence that structured spatial understanding is difficult not only for end-to-end action generation, but also for explicit target grounding and target-to-action conversion pipelines.
\begin{table}[t]
\centering
\caption{
Information access for end-to-end VLA policies and modular baselines. 
Both method families are evaluated without privileged simulator information during inference. 
Modular baselines use explicit VLM-based grounding and scripted execution, so their results are interpreted as cross-paradigm comparisons with additional diagnostic value.
}
\label{tab:info_access}
\small\setlength{\tabcolsep}{4pt}
\begin{tabular}{lcc}
\toprule
\textbf{Information} & \textbf{VLA policies} & \textbf{Modular baselines} \\
\midrule
Language instruction & Yes & Yes \\
RGB observations & Yes & Yes \\
Robot proprioception & Yes & Yes \\
Object poses & No & No \\
Object identities & No & No \\
Segmentation masks & No & No \\
Task metadata & No & No \\
Oracle target coordinates & No & No \\
Ground-truth task variables & No & No \\
\geomind-specific fine-tuning & No & No \\
\midrule
Role in evaluation & End-to-end policy & Modular baseline \\
Inference structure & Direct action prediction & VLM grounding + scripted execution \\
Diagnostic signal & Trajectory/action behavior & Grounding and target-selection errors \\
\bottomrule
\end{tabular}
\end{table}

\subsection{Fairness and Information Access}
\label{app:fairness}

The main comparison in GEOMIND covers multiple embodied-agent paradigms under the same task instances, random seeds, workspace constraints, episode horizon, and success criteria. 
For end-to-end VLA policies, all models receive the same observation modalities and are evaluated as frozen policies without \geomind-specific fine-tuning. 
For modular baselines, the VLM grounding stage receives the same task instruction and RGB observations, while execution is performed by a scripted controller after target selection.

No evaluated method receives privileged simulator information during inference. 
Object poses, simulator-resolved object identities, oracle target coordinates, ground-truth task variables, segmentation masks, and task metadata are excluded from both VLA policies and modular baselines. 
These states are used only for evaluation, success checking, and diagnostic logging.

Since VLA policies and modular baselines differ in inference structure, their results should be read as a cross-paradigm comparison rather than a comparison of identical policy interfaces. 
This is intentional: GEOMIND aims to evaluate whether different embodied-agent designs can solve the same structured spatial objectives under the same task conditions. 
The modular results complement the VLA results by revealing whether explicit visual grounding and scripted execution can overcome the spatial failures observed in end-to-end policies.

\section{Spatial Structure Formalization}
\label{app:spatial_structure_formalization}

\begin{figure}[t]
\centering
\includegraphics[width=\linewidth]{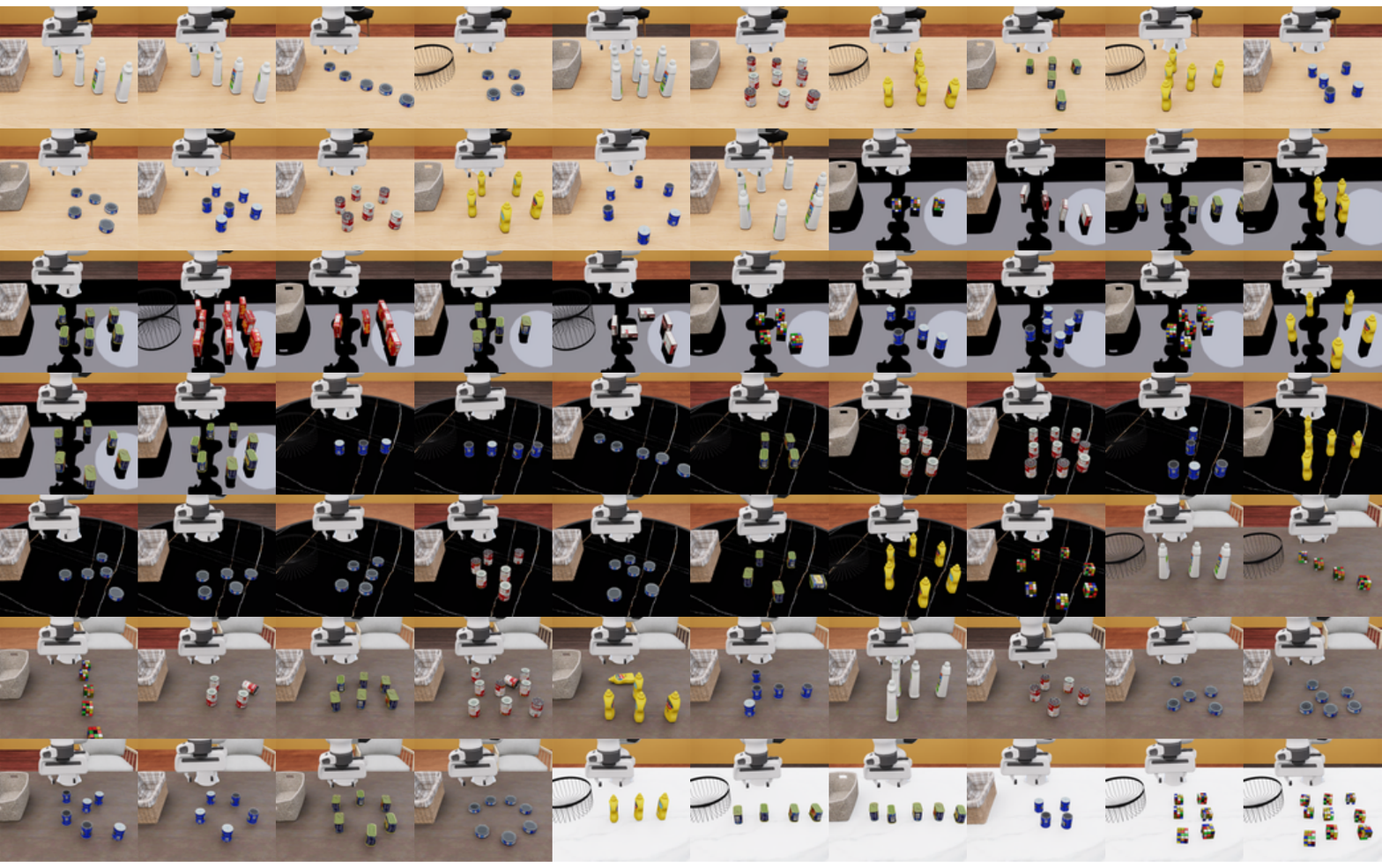}
\vspace{-1.5em}
\caption{Example layouts generated by the spatial structure modeling and construction pipeline.} 
\vspace{-1.5em}
\label{fig:spatial_structure_gallery}
\end{figure}

This appendix gives a compact description of the spatial structures used in \geomind.
Fig.~\ref{fig:spatial_structure_gallery} shows representative structure layouts.
Our goal is not to bind each task to one fixed object coordinate template.
Instead, each task is associated with a spatial structure, and \geomind samples multiple physical realizations that preserve the same task-relevant relation.
\subsection{Structure-Preserving Sampling}
\label{app:structure_sampling}

Let a scene contain $N_{\mathrm{obj}}$ objects \(\mathcal{O}=\{o_i\}_{i=1}^{N_{\mathrm{obj}}}\).
The pose of object \(o_i\) is
\begin{equation}
\mathbf{x}_i=(\mathbf{p}_i,\mathbf{q}_i),
\end{equation}
where \(\mathbf{p}_i\) is the object position and \(\mathbf{q}_i\) is the object orientation.
For tabletop relations, we use the planar position \(\bar{\mathbf{p}}_i=(x_i,y_i)\).
The full scene state is
\begin{equation}
\mathbf{X}=\{\mathbf{x}_i\}_{i=1}^{N_{\mathrm{obj}}}.
\end{equation}

We denote a spatial structure as \(S\).
It specifies what should remain invariant during sampling, such as an object order, a grid layout, an anchor-based relation, or a geometric target rule.
Given \(S\), \geomind samples object poses as
\begin{equation}
\mathbf{X}\sim\mathcal{P}(\mathbf{X}\mid S,\epsilon),
\end{equation}
where \(\epsilon\) controls small layout variations such as jitter or spacing changes.
A sampled scene is accepted only if
\begin{equation}
\mathbf{X}\in\Omega_{\mathrm{struct}}(S,\epsilon)\cap\Omega_{\mathrm{phys}}.
\end{equation}
Here, $\Omega_{\mathrm{struct}}(S,\epsilon)$ is the set of configurations satisfying the structural constraints, and $\Omega_{\mathrm{phys}}$ is the set of physically valid configurations.

In short, \(S\) describes the stable spatial meaning of the task, while \(\mathbf{X}\) is one sampled scene that realizes this meaning.

\subsection{Representative Spatial Structures}
\label{app:representative_spatial_structures}

As illustrated in Fig.~\ref{fig:spatial_structure_gallery}, \geomind uses several common spatial structures.
We summarize them below with minimal constraints.

\paragraph{Ordered structure.}
This structure is used for tasks such as ``pick the leftmost object'' or ``pick the second object from the right.''
Let \(\mathbf{u}\) be a unit vector in the ordering direction and define the projected coordinate
\begin{equation}
s_i=\mathbf{u}^{\top}\bar{\mathbf{p}}_i.
\end{equation}
Let $\pi$ be the permutation specifying the intended object order $o_{\pi(1)}\prec\cdots\prec o_{\pi(N_{\mathrm{obj}})}$. For a minimum projected separation $\delta_{\min}>0$, a valid sample must satisfy
\begin{equation}
s_{\pi(k+1)}-s_{\pi(k)}\geq\delta_{\min},\qquad k=1,\ldots,N_{\mathrm{obj}}-1.
\end{equation}
This keeps the object order unchanged while allowing the exact coordinates to vary.

\paragraph{Grid structure.}
This structure is used for row-column layouts, such as ``the object in the top-left cell'' or ``the object in the same row as the bowl.''
Each object has a grid index \(g_i=(r_i,c_i)\).
A sampled position is generated around the corresponding grid cell:
\begin{equation}
\bar{\mathbf{p}}_i
=
\mathbf{c}
+(c_i-1)d_x\mathbf{u}
+(r_i-1)d_y\mathbf{v}
+\boldsymbol{\eta}_i,
\qquad
\|\boldsymbol{\eta}_i\|_{\infty}\leq\epsilon.
\end{equation}
Here, $\mathbf{c}$ is the grid origin, $d_x,d_y>0$ are the grid spacings, $\mathbf{u}$ and $\mathbf{v}$ are unit vectors along the column and row directions, and $\boldsymbol{\eta}_i$ is the positional perturbation. The perturbation bound alone does not guarantee that row-column relations are preserved; accepted configurations must also satisfy the structural-validity condition above.

\paragraph{Anchor-based relation.}
This structure is used for tasks such as ``the object left of the cup'', ``the object near the plate'', or ``the object between two bowls.''
For an object \(o_i\) and an anchor \(o_a\), define
\begin{equation}
\Delta_{ia}=\bar{\mathbf{p}}_i-\bar{\mathbf{p}}_a.
\end{equation}
A directional relation is valid when
\begin{equation}
\mathbf{u}_{r}^{\top}\Delta_{ia}\geq\tau_r,
\end{equation}
where $\mathbf{u}_r$ is a unit vector in the relation direction and $\tau_r>0$ is a directional separation threshold that excludes ambiguous cases. Let $R(o_i,o_a)\in\{0,1\}$ be the corresponding task predicate over candidate objects and the anchor; ineligible objects have predicate value zero.
The target is accepted only when the relation selects one unique object:
\begin{equation}
\{o_i\in\mathcal{O}:R(o_i,o_a)=1\}=\{o^*\}.
\end{equation}

\paragraph{Geometric target.}
This structure is used when the target is an implicit position rather than a visible object.
For example, the target may be the midpoint between two anchors:
\begin{equation}
\mathbf{g}_{\mathrm{mid}}
=
\frac{1}{2}(\bar{\mathbf{p}}_a+\bar{\mathbf{p}}_b),
\end{equation}
or the centroid of the positions of a nonempty set of anchors indexed by $\mathcal{A}$:
\begin{equation}
\mathbf{g}_{\mathrm{center}}
=
\frac{1}{|\mathcal{A}|}
\sum_{a\in\mathcal{A}}\bar{\mathbf{p}}_a.
\end{equation}
These tasks require the agent to infer a target from scene geometry, instead of directly selecting an existing object.

\subsection{Dynamic Relations}
\label{app:dynamic_relations}

For tasks with scene changes, \geomind distinguishes two cases.
In scene-updating tasks, the target is determined by the current relation:
\begin{equation}
\{o_i\in\mathcal{O}:R_t(o_i,o_a)=1\}=\{o_t^*\}.
\end{equation}
For example, the correct object is the one currently left of the cup.

In identity-tracking tasks, the target is determined by the initial relation and then kept fixed:
\begin{equation}
\{o_i\in\mathcal{O}:R_0(o_i,o_a)=1\}=\{o^*\},
\qquad
\mathrm{target}(t)=o^*.
\end{equation}
For example, the correct object is the one that was originally left of the cup, even if it later moves away.

\subsection{Physical Validation}
\label{app:physical_validation}

Finally, each sampled scene is checked in the simulator.
We retain instances that satisfy the simulator's collision and workspace checks and have a reachable target. For distinct objects whose footprints must remain separated in the tabletop plane, a circular-footprint separation condition can be written as
\begin{equation}
\|\bar{\mathbf{p}}_i-\bar{\mathbf{p}}_j\|_2
\geq
\rho_i+\rho_j+d_{\mathrm{safe}},
\qquad
\bar{\mathbf{p}}_i\in\Omega_{\mathrm{table}},
\qquad
\mathbf{g}\in\Omega_{\mathrm{reach}}.
\end{equation}
Here, $\rho_i$ is the radius of a bounding disk for object $i$'s planar footprint, $d_{\mathrm{safe}}\geq0$ is a separation margin, $\Omega_{\mathrm{table}}$ is the admissible tabletop region, and $\mathbf{g}\in\Omega_{\mathrm{reach}}$ specifies a reachable goal position. The pairwise inequality applies only to distinct object pairs requiring planar separation; it is not a universal constraint for containment or stacking. Task-specific contact feasibility is checked in simulation.

\subsection{Summary}
\label{app:structure_summary}

The structure gallery in Fig.~\ref{fig:spatial_structure_gallery} provides visual examples of the layouts used in \geomind.
The formulas above specify only the minimal constraints needed to preserve these structures during sampling.
This allows \geomind to generate diverse scenes with the same task semantics, so evaluation focuses on spatial understanding rather than memorization of fixed coordinates.

\Needspace{9\baselineskip}
\section{Task Details}
\label{app:task_details}

We present 26 illustrative task cards covering a subset of the benchmark. The full evaluation set contains 54 task variants: 20 Base, 7 Semantic, 7 Visual, and 20 Execution variants. The latter three categories constitute the 34 augmented variants. The number of task cards is not the total number of evaluated variants.

Each card gives a compact specification of one task variant, including the task instruction, the corresponding suite and task family, the spatial structure used for scene generation, the ground-truth target, the numerical ground-truth pose, and the evaluation criterion.
The scene image on the left illustrates one sampled realization of the task, while the fields on the right define the task in a reproducible form.
All target poses are reported in the simulator world frame, and distances are measured in meters unless otherwise stated.

\definecolor{TaskCardLine}{HTML}{D8D8E8}
\definecolor{TaskCardTitle}{HTML}{333333}

\newcommand{\geomindtaskcard}[9]{%
\Needspace{4.3cm}%
\vspace{0.65em}
\noindent
\begingroup
\textcolor{TaskCardLine}{\rule{\linewidth}{0.4pt}}

\vspace{0.35em}

\begin{minipage}{0.98\linewidth}
\begin{minipage}[t]{0.30\linewidth}
\vspace{0pt}
\centering
\includegraphics[width=\linewidth,height=3cm,keepaspectratio]{#4}
\end{minipage}
\hfill
\begin{minipage}[t]{0.65\linewidth}
\vspace{0pt}
\scriptsize

{\small\textbf{\textcolor{TaskCardTitle}{#1}}}

\vspace{0.30em}

\textbf{Task suite and family.} #2 / #3

\vspace{0.22em}
\textbf{Instruction.} #5

\vspace{0.22em}
\textbf{Structure.} #6

\vspace{0.22em}
\textbf{GT Target.} #7

\vspace{0.22em}
\textbf{GT Pose.} #8

\vspace{0.22em}
\textbf{Eval.} #9

\end{minipage}
\end{minipage}

\endgroup
}


\geomindtaskcard
{BindingBase\_AnchorDirection\_v0}
{\geomindbase}
{Target Identification}
{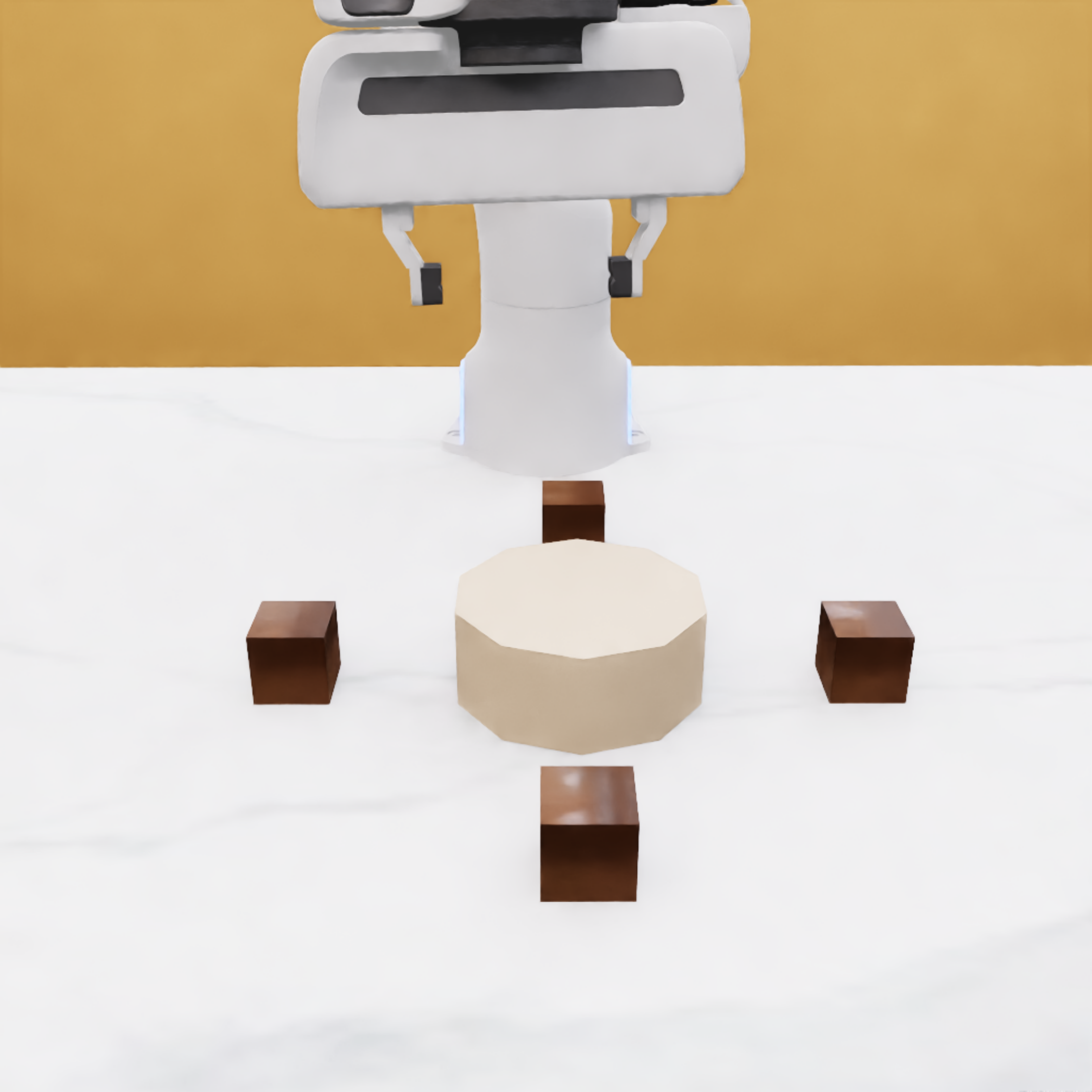}
{Move the end effector near the cube to the right of the blue bowl.}
{anchor\_direction\_injected structure.}
{Protocol target object block\_right proximity region.}
{$\mathbf{p}^*=(0.40, -0.15, 0.03)$ m; $\theta^*=0.00$ rad.}
{Success requires the end-effector distance to the target object's center to be $< 0.06$ m and the final state to remain stable for 1.0 s.}

\geomindtaskcard
{ConstraintBase\_SelectExclusionUnique\_v0}
{\geomindbase}
{Constraint Composition}
{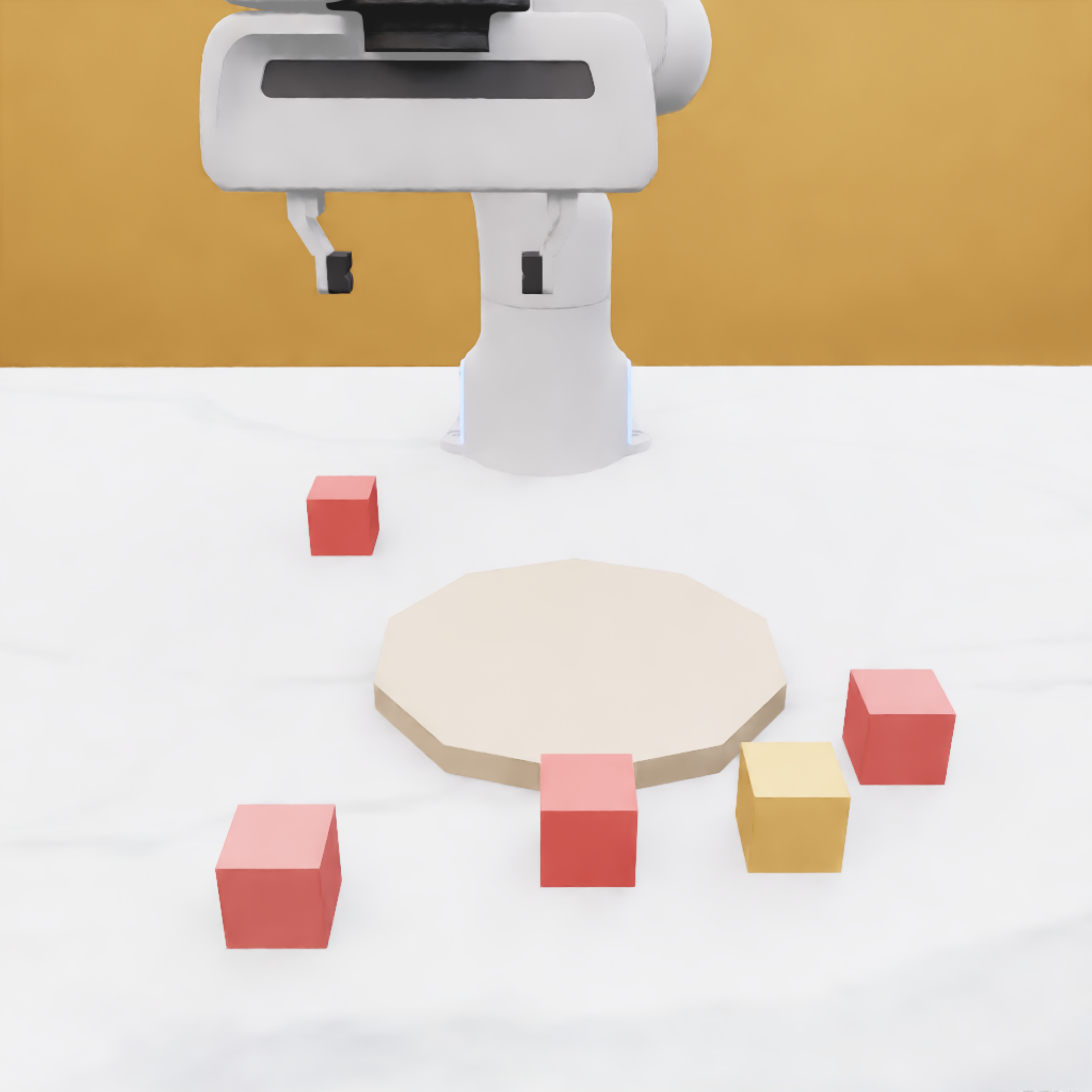}
{Move the end effector near the red cube that is closest to the blue plate and is not the leftmost cube.}
{exclusion\_distance\_constraint\_injected structure.}
{Protocol target object red\_block\_target proximity region.}
{$\mathbf{p}^*=(0.56, 0.02, 0.03)$ m; $\theta^*=0.00$ rad.}
{Success requires the end-effector distance to the target object's center to be $< 0.06$ m and the final state to remain stable for 1.0 s.}

\geomindtaskcard
{ConstraintBase\_SelectVisualSpatial\_v0}
{\geomindbase}
{Constraint Composition}
{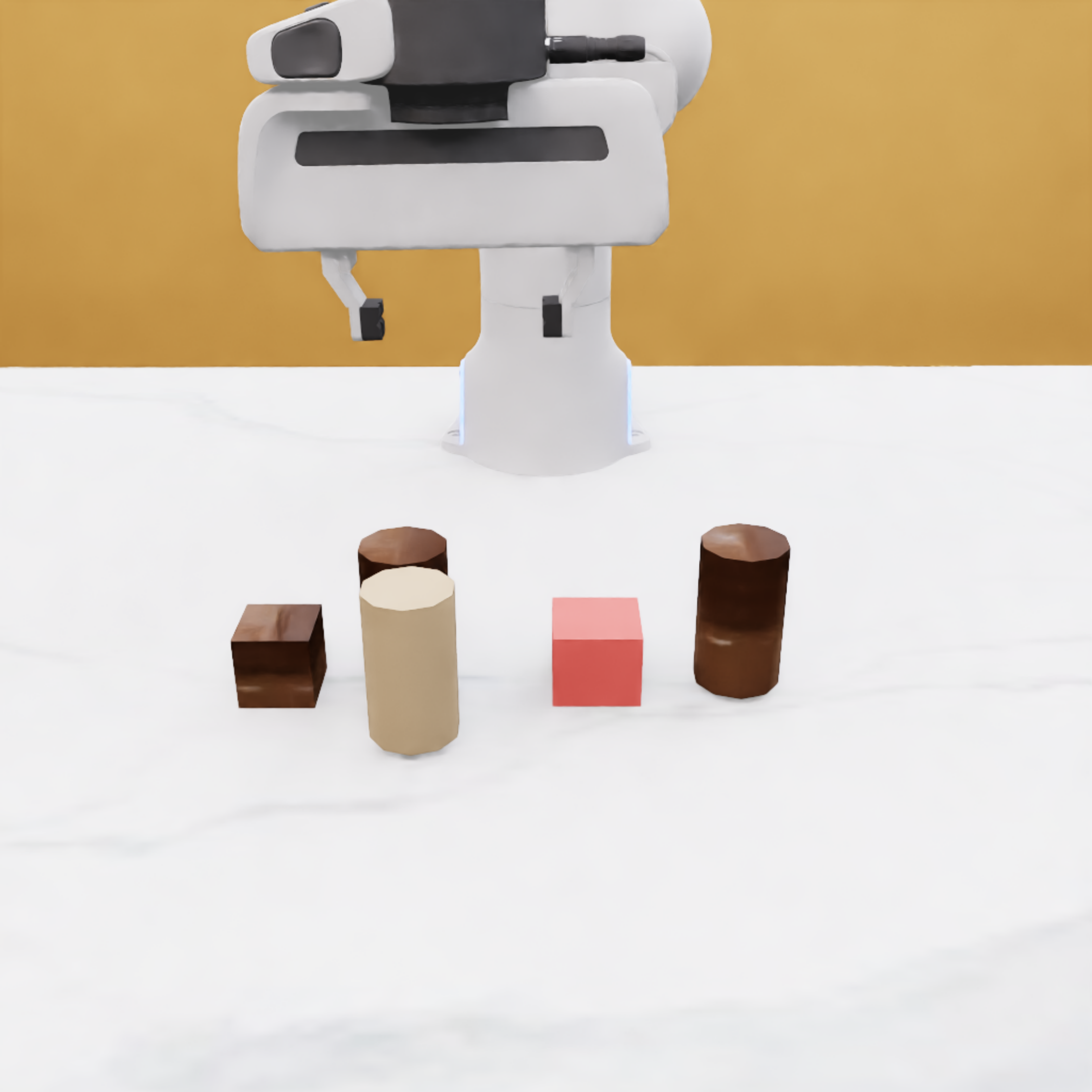}
{Move the end effector near the wood-grain cylinder to the right of the red block.}
{visual\_spatial\_constraint\_injected structure.}
{Protocol target object wood\_cylinder\_target proximity region.}
{$\mathbf{p}^*=(0.40, -0.08, 0.03)$ m; $\theta^*=0.00$ rad.}
{Success requires the end-effector distance to the target object's center to be $< 0.06$ m and the final state to remain stable for 1.0 s.}

\geomindtaskcard
{GeometricBase\_IsometricExtension\_v0}
{\geomindbase}
{Geometric Inference}
{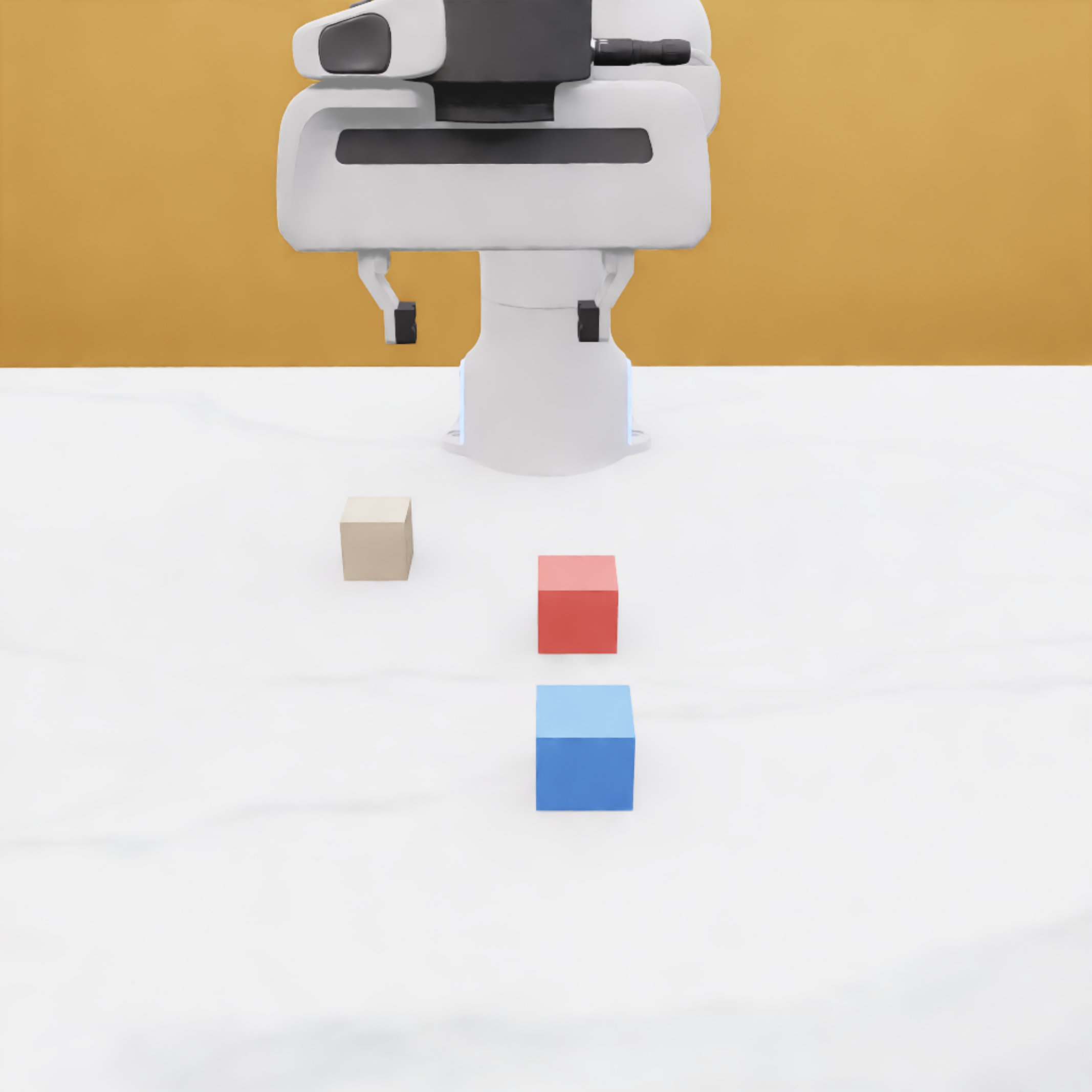}
{You already hold the cube. Place it on the red–blue line extension at equal spacing.}
{2-anchor line\_blocks\_runtime\_injected structure.}
{Implicit equal-spacing extension point beyond the second anchor.}
{$\mathbf{p}^*=(0.66, 0.02, 0.03)$ m; $\theta^*=0.00$ rad.}
{Success requires the distance from the object center to the target position to be $< 0.05$ m and the final state to remain stable for 1.0 s.}

\geomindtaskcard
{GeometricBase\_PerfectMidpoint\_v0}
{\geomindbase}
{Geometric Inference}
{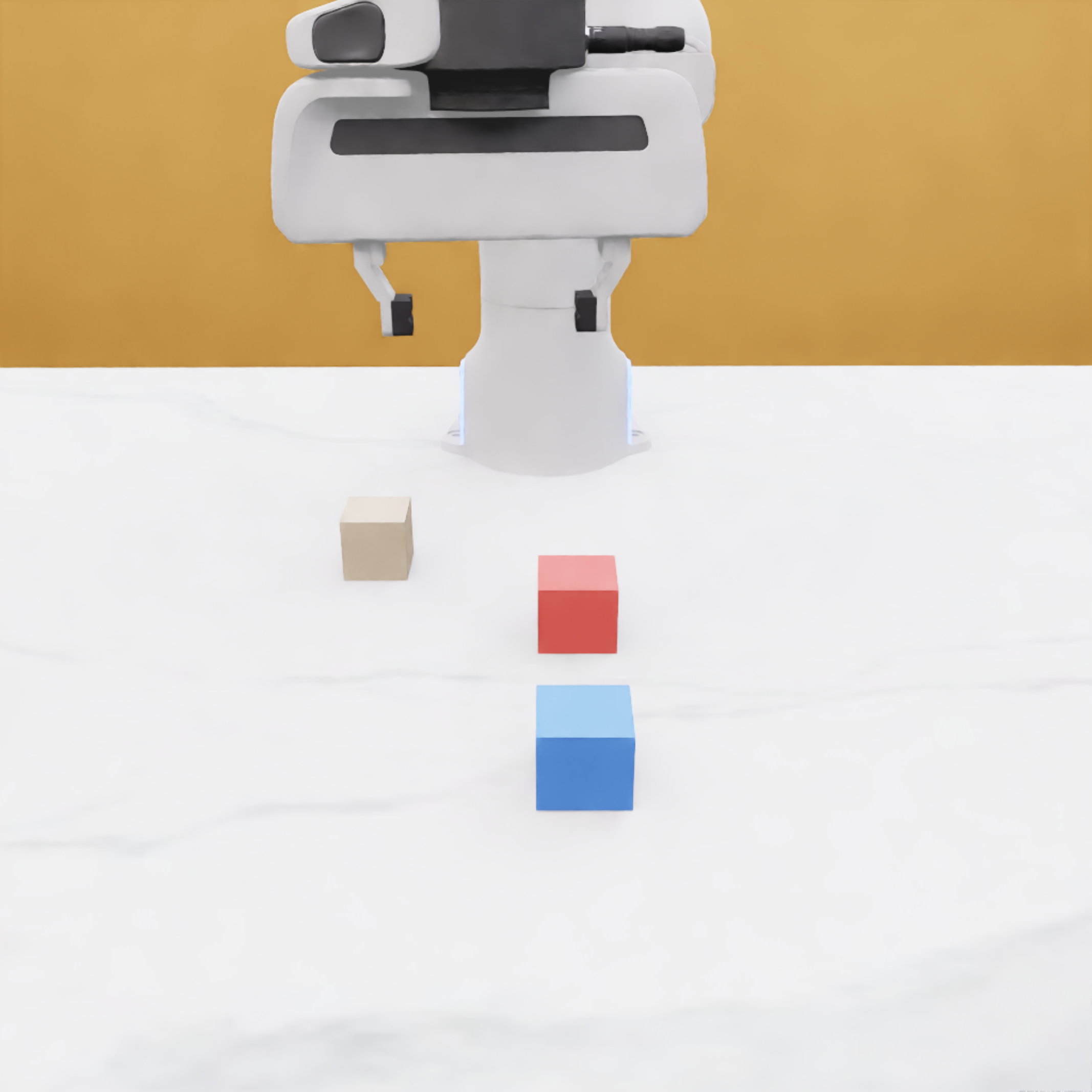}
{You already hold the cube. Place it midway between the red and blue reference cubes.}
{2-anchor line\_blocks\_runtime\_injected structure.}
{Implicit midpoint region between the two colored anchor blocks.}
{$\mathbf{p}^*=(0.42, 0.02, 0.03)$ m; $\theta^*=0.00$ rad.}
{Success requires the distance from the object center to the target position to be $< 0.05$ m and the final state to remain stable for 1.0 s.}

\geomindtaskcard
{GeometricBase\_PointSymmetry\_v0}
{\geomindbase}
{Geometric Inference}
{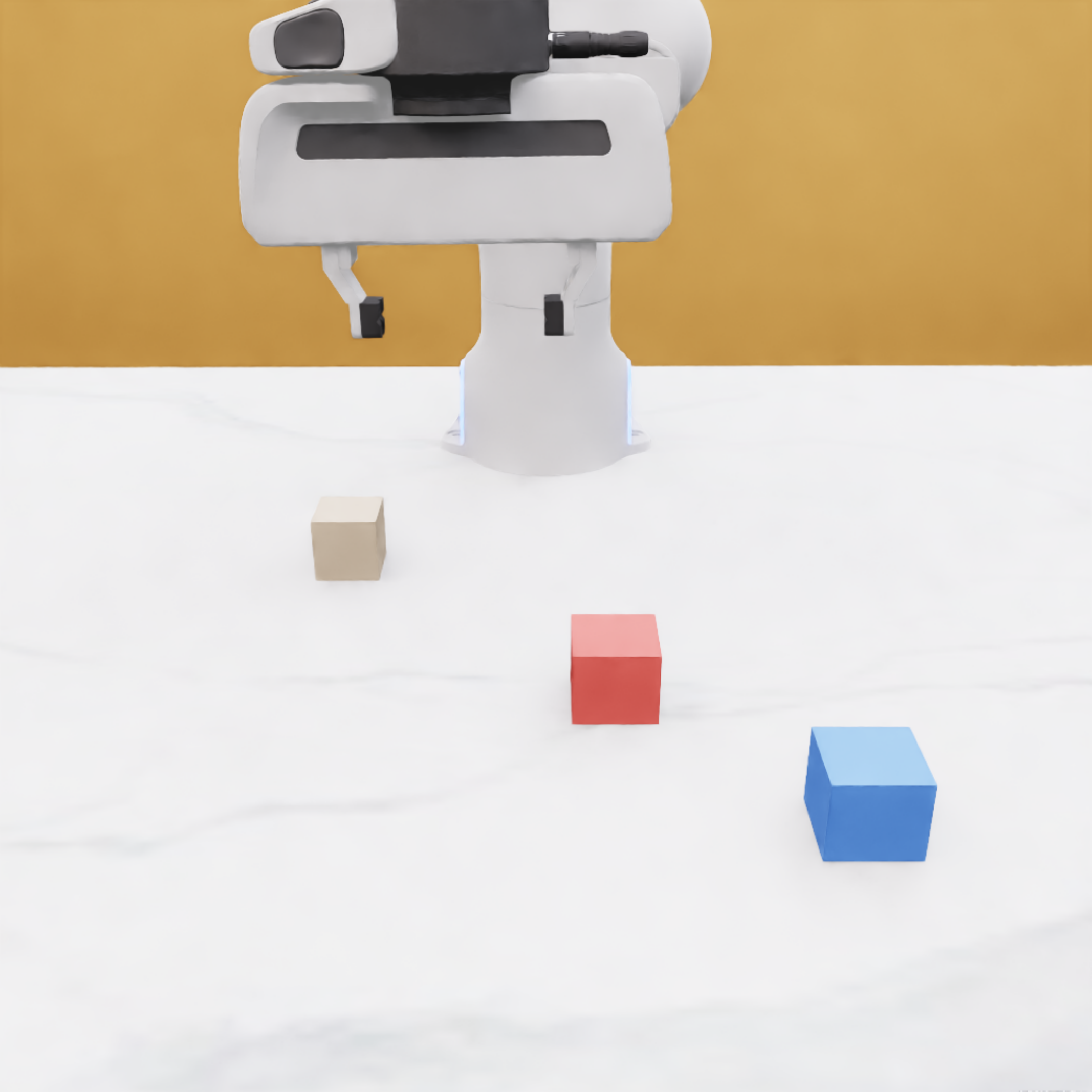}
{You already hold the cube. Place it symmetrically to the blue cube about the red cube.}
{2-anchor diagonal\_blocks\_runtime\_injected structure.}
{Implicit point-symmetric reflection of the reference anchor across the center anchor.}
{$\mathbf{p}^*=(0.30, -0.08, 0.03)$ m; $\theta^*=0.00$ rad.}
{Success requires the distance from the object center to the target position to be $< 0.05$ m and the final state to remain stable for 1.0 s.}

\geomindtaskcard
{GeometricBase\_SquareCompletion\_v0}
{\geomindbase}
{Geometric Inference}
{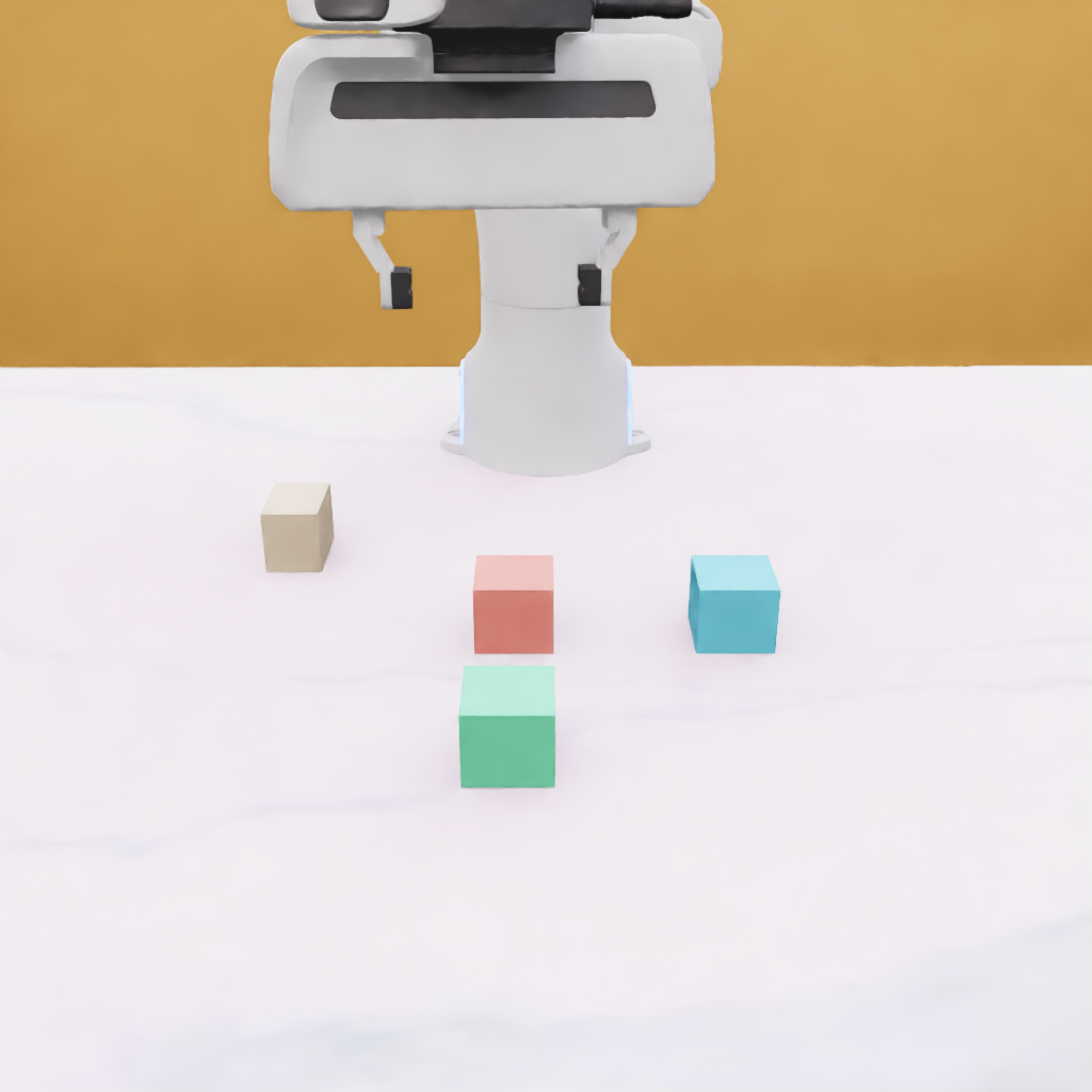}
{You already hold the fourth cube. Place it to complete a square with the three anchor cubes.}
{3-anchor square\_corner\_runtime\_injected structure.}
{Implicit missing square corner inferred from three anchors.}
{$\mathbf{p}^*=(0.48, 0.12, 0.03)$ m; $\theta^*=0.00$ rad.}
{Success requires the distance from the object center to the target position to be $< 0.05$ m and the final state to remain stable for 1.0 s.}

\geomindtaskcard
{GeometricBase\_TrendContinuation\_v0}
{\geomindbase}
{Geometric Inference}
{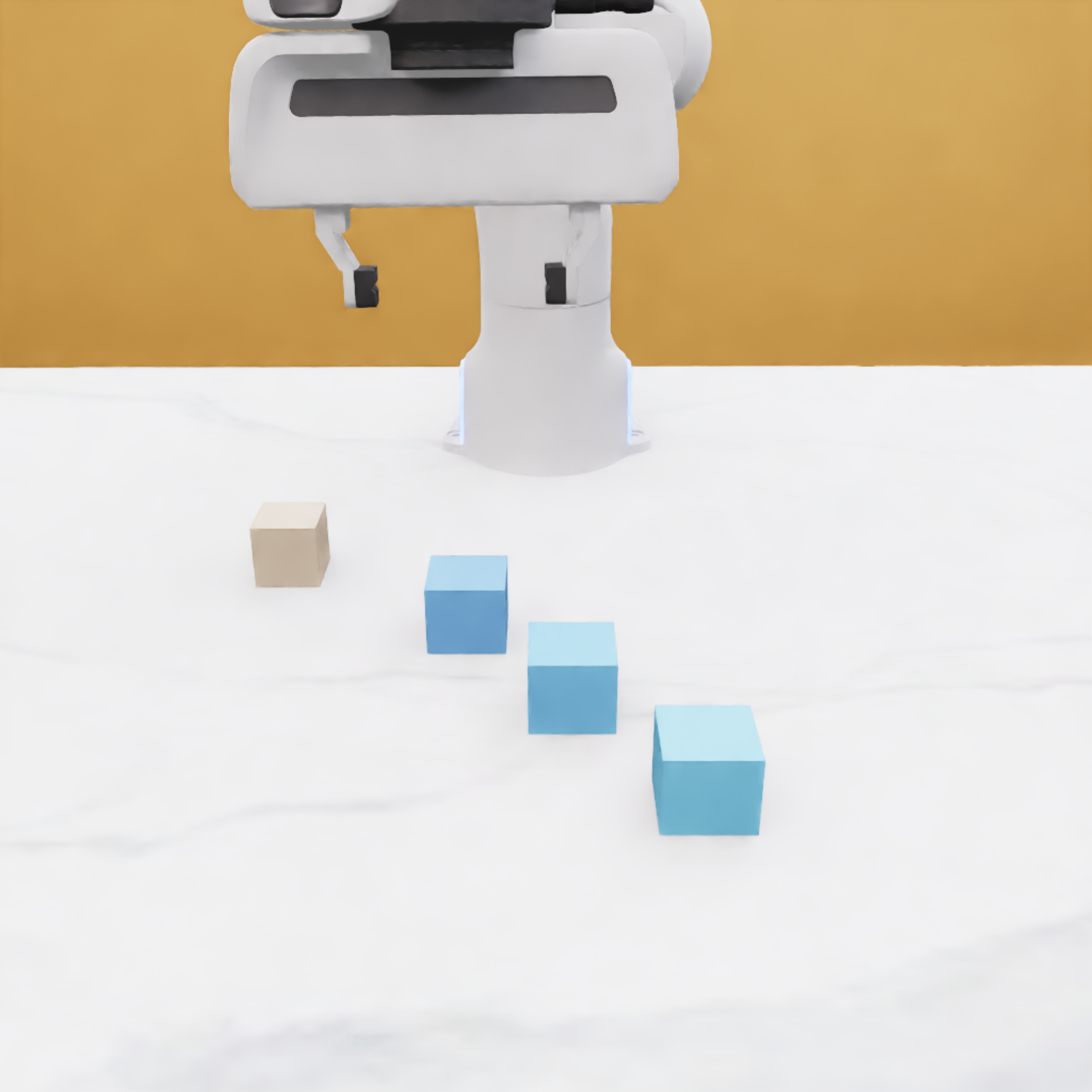}
{You already hold the next cube. Place it to continue the visible row/column pattern.}
{3-anchor diagonal\_trend\_blocks\_runtime\_injected structure.}
{Implicit next point in the anchor trend.}
{$\mathbf{p}^*=(0.61, 0.15, 0.03)$ m; $\theta^*=0.00$ rad.}
{Success requires the distance from the object center to the target position to be $< 0.05$ m and the final state to remain stable for 1.0 s.}

\geomindtaskcard
{BindingBase\_GridRowcolGrounding\_v0}
{\geomindbase}
{Target Identification}
{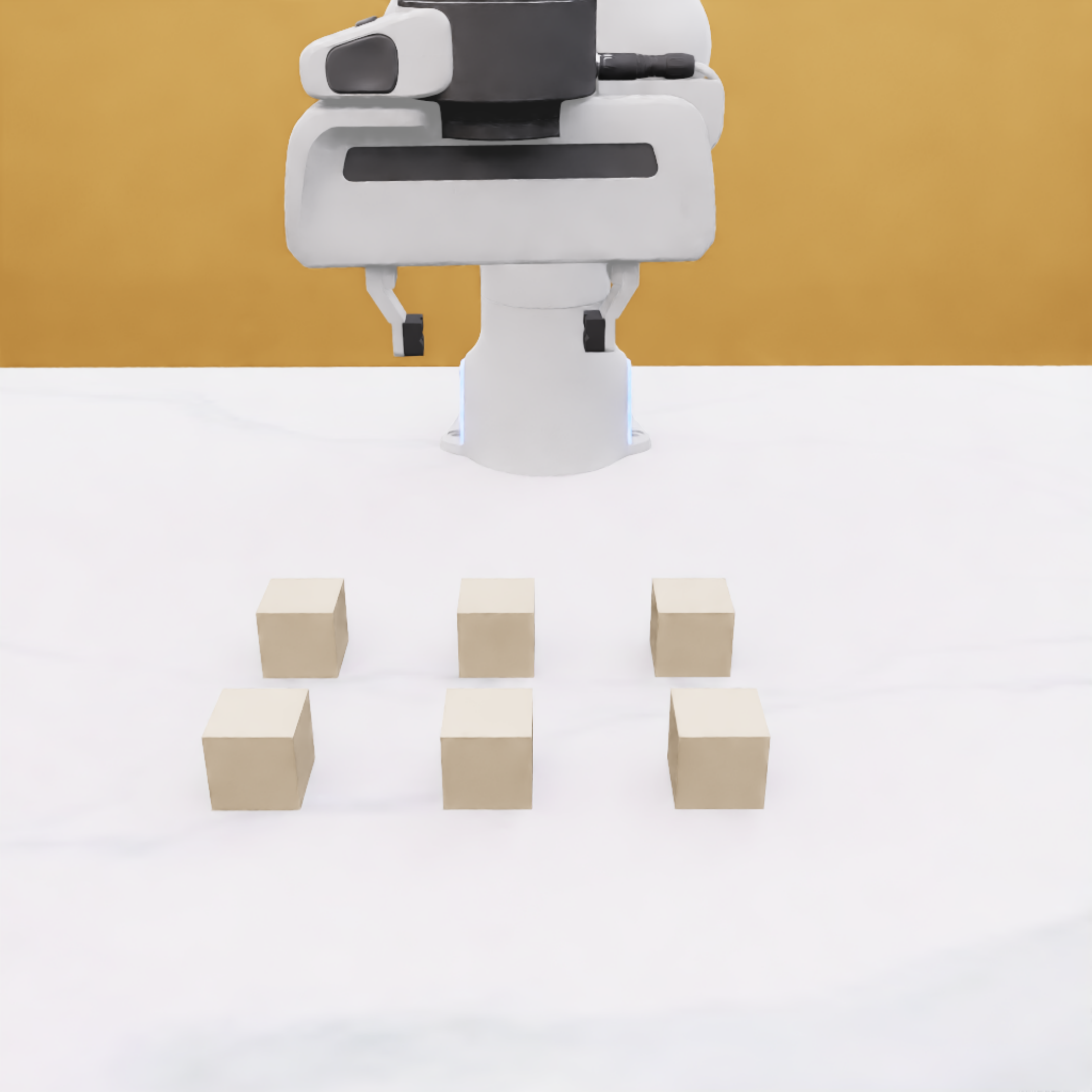}
{Move the end effector near the cube at row 1, column 2.}
{runtime\_grid\_2x3\_static structure.}
{Protocol target object grid\_cell\_1 proximity region.}
{$\mathbf{p}^*=(0.37, -0.03, 0.03)$ m; $\theta^*=0.00$ rad.}
{Success requires the end-effector distance to the target object's center to be $< 0.06$ m and the final state to remain stable for 1.0 s.}

\geomindtaskcard
{BindingBase\_OrdinalInsertStatic\_v0}
{\geomindbase}
{Target Identification}
{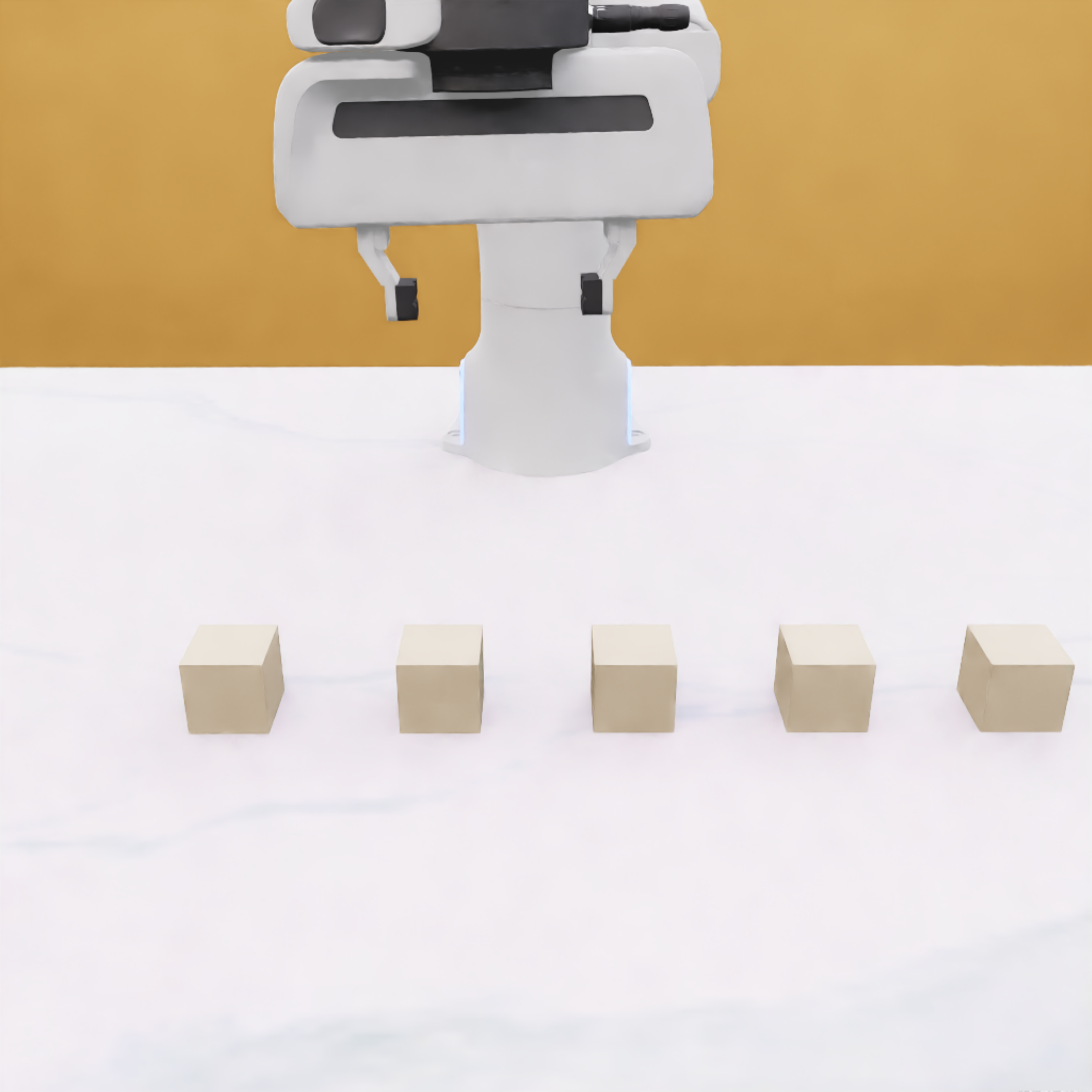}
{Move the end effector near the second cube from the left.}
{runtime\_block\_line\_static\_ordinal structure.}
{Protocol target object ordinal\_static\_1 proximity region.}
{$\mathbf{p}^*=(0.43, -0.06, 0.03)$ m; $\theta^*=0.00$ rad.}
{Success requires the end-effector distance to the target object's center to be $< 0.06$ m and the final state to remain stable for 1.0 s.}

\geomindtaskcard
{ReindexBase\_ReindexExtreme\_v0}
{\geomindbase}
{Scene Updating}
{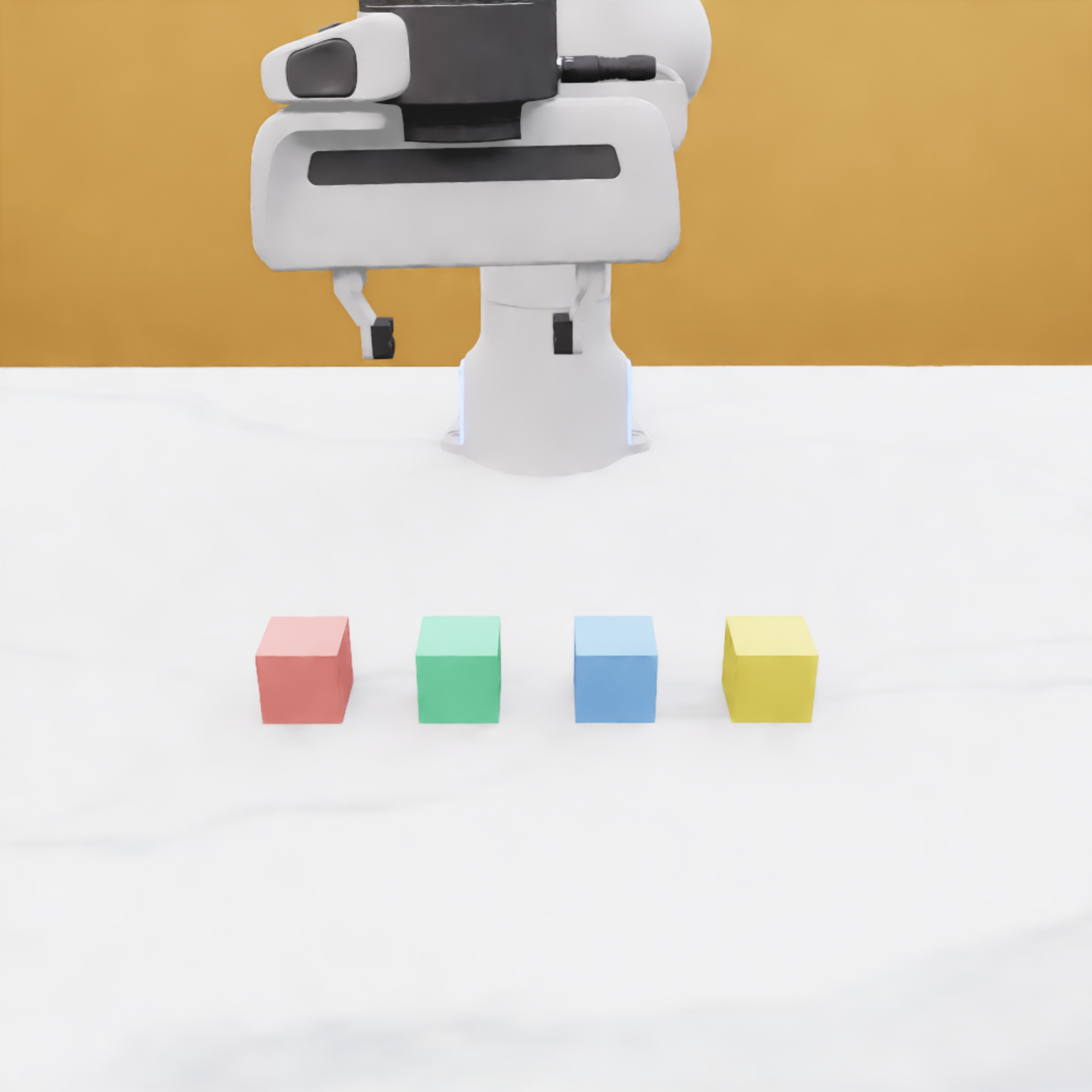}
{Move the end effector near the leftmost cube after reordering.}
{runtime\_block\_line structure.}
{Protocol target object reindex\_block\_1 proximity region.}
{$\mathbf{p}^*=(0.42, -0.05, 0.03)$ m; $\theta^*=0.00$ rad.}
{Success requires the end-effector distance to the target object's center to be $< 0.06$ m and the final state to remain stable for 1.0 s.}

\geomindtaskcard
{ReindexBase\_ReindexInsert\_v0}
{\geomindbase}
{Scene Updating}
{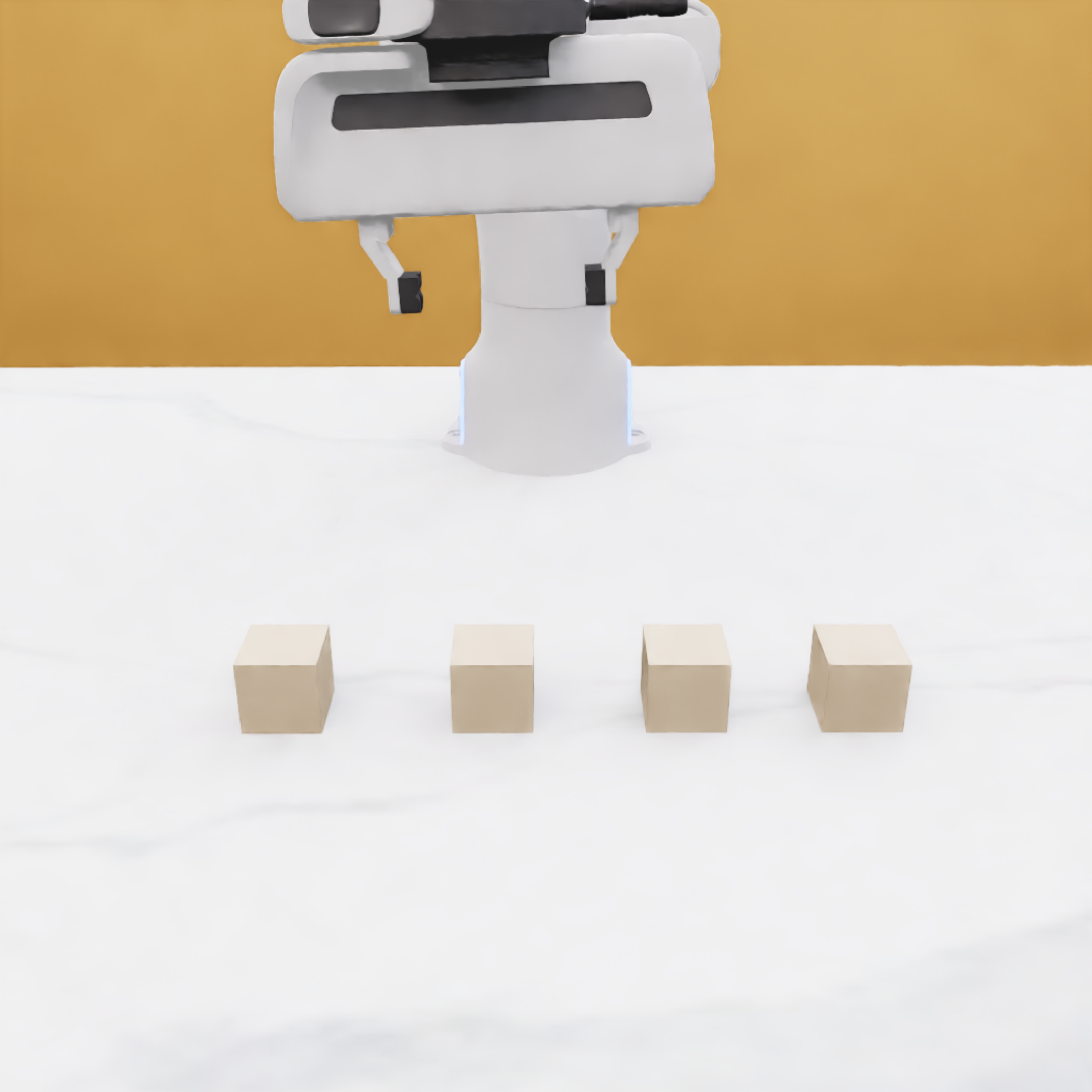}
{Move the end effector near the second-from-left cube.}
{runtime\_block\_line\_insert structure.}
{Protocol target object insert\_block\_3 proximity region.}
{$\mathbf{p}^*=(0.43, 0.18, 0.03)$ m; $\theta^*=0.00$ rad.}
{Success requires the end-effector distance to the target object's center to be $< 0.06$ m and the final state to remain stable for 1.0 s.}

\geomindtaskcard
{ReindexBase\_ReindexNeighbor\_v0}
{\geomindbase}
{Scene Updating}
{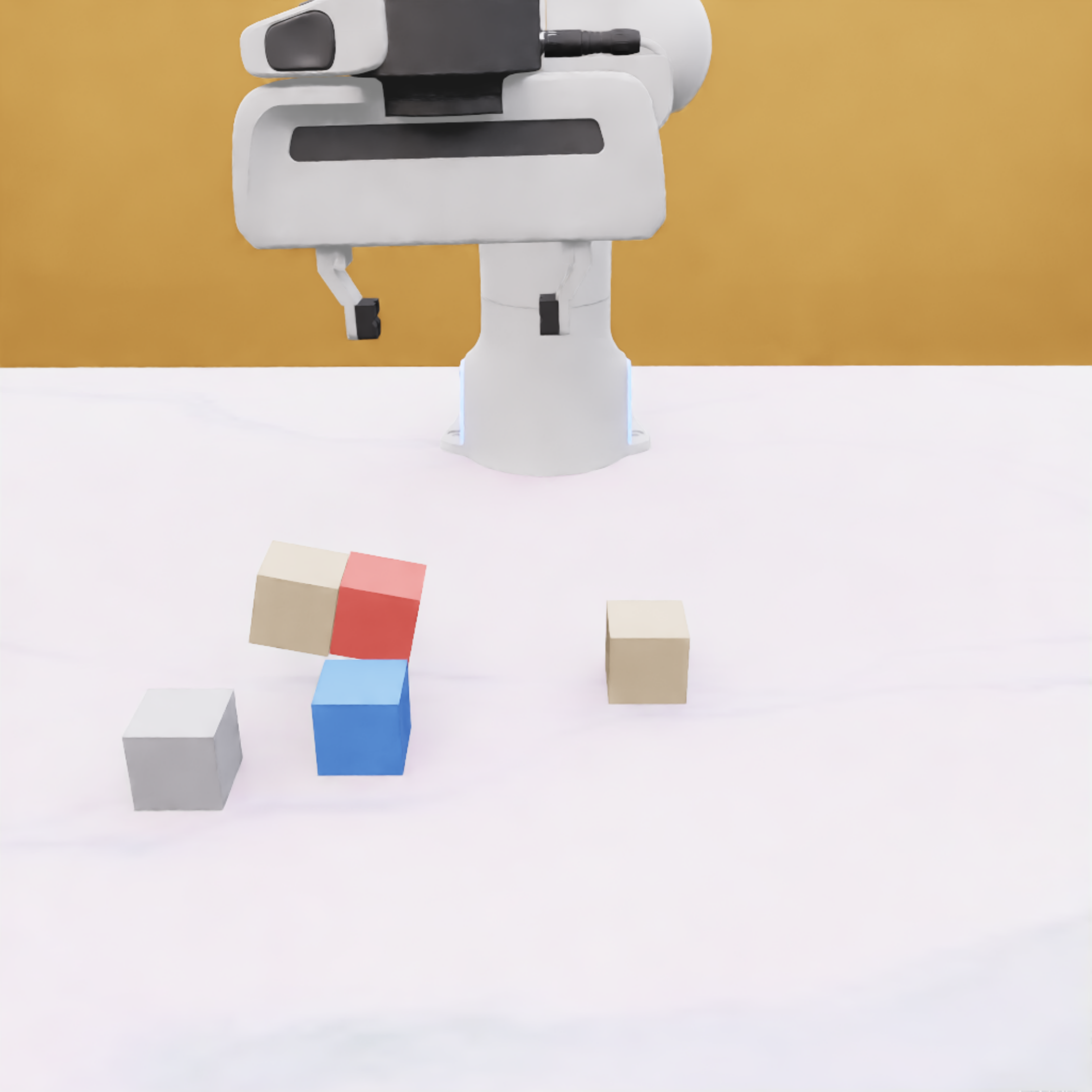}
{Move the end effector near the cube immediately left of the red cube.}
{runtime\_red\_anchor\_neighbor structure.}
{Protocol target object right\_neighbor\_block proximity region.}
{$\mathbf{p}^*=(0.40, 0.06, 0.03)$ m; $\theta^*=0.00$ rad.}
{Success requires the end-effector distance to the target object's center to be $< 0.06$ m and the final state to remain stable for 1.0 s.}

\geomindtaskcard
{ReindexBase\_ReindexSubset\_v0}
{\geomindbase}
{Scene Updating}
{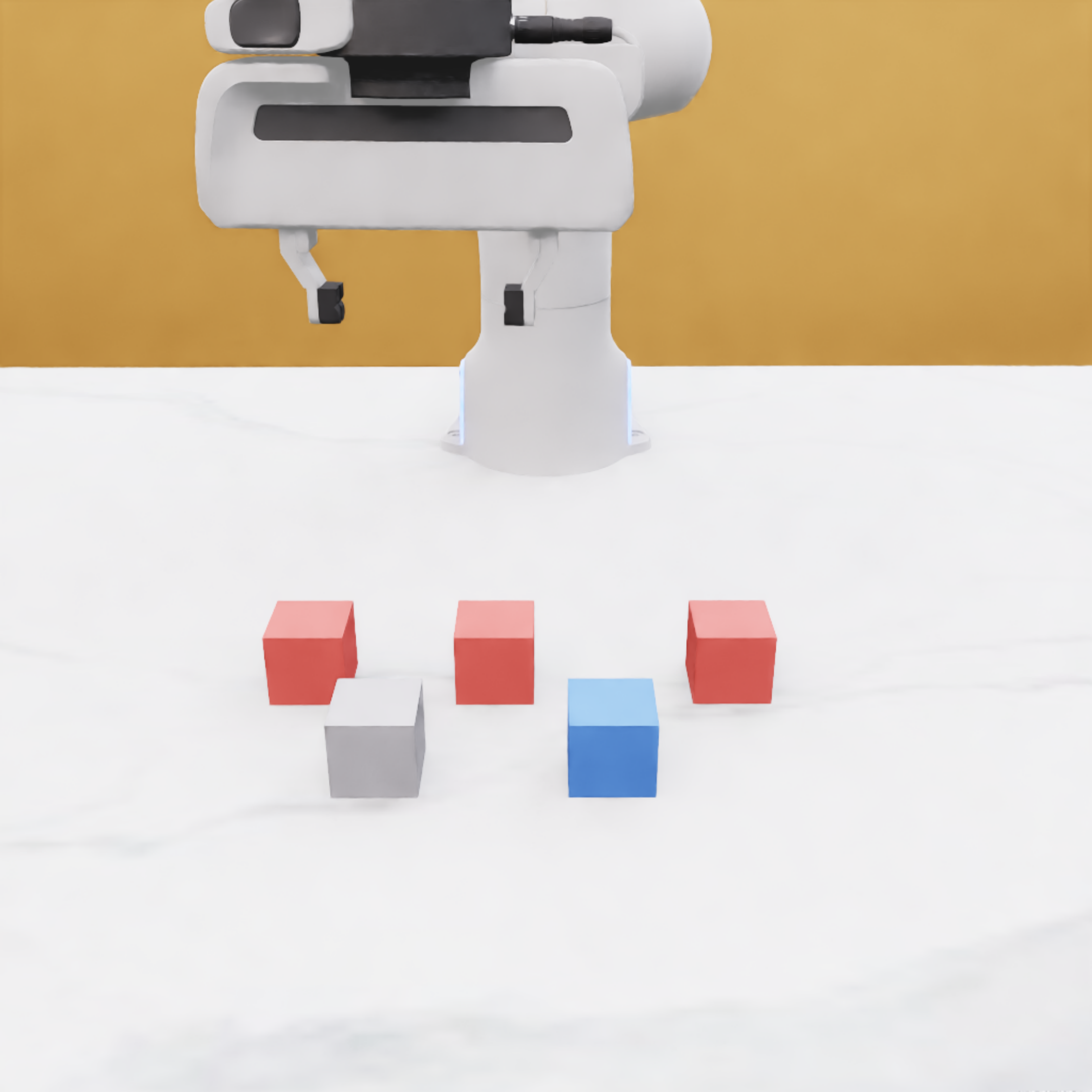}
{Move the end effector near the middle cube among the red cubes only.}
{runtime\_red\_subset\_with\_distractors structure.}
{Protocol target object red\_subset\_block\_2 proximity region.}
{$\mathbf{p}^*=(0.40, 0.11, 0.03)$ m; $\theta^*=0.00$ rad.}
{Success requires the end-effector distance to the target object's center to be $< 0.06$ m and the final state to remain stable for 1.0 s.}

\geomindtaskcard
{ReindexAugSem\_ReindexExtreme\_v0}
{\geomindaug}
{Scene Updating}
{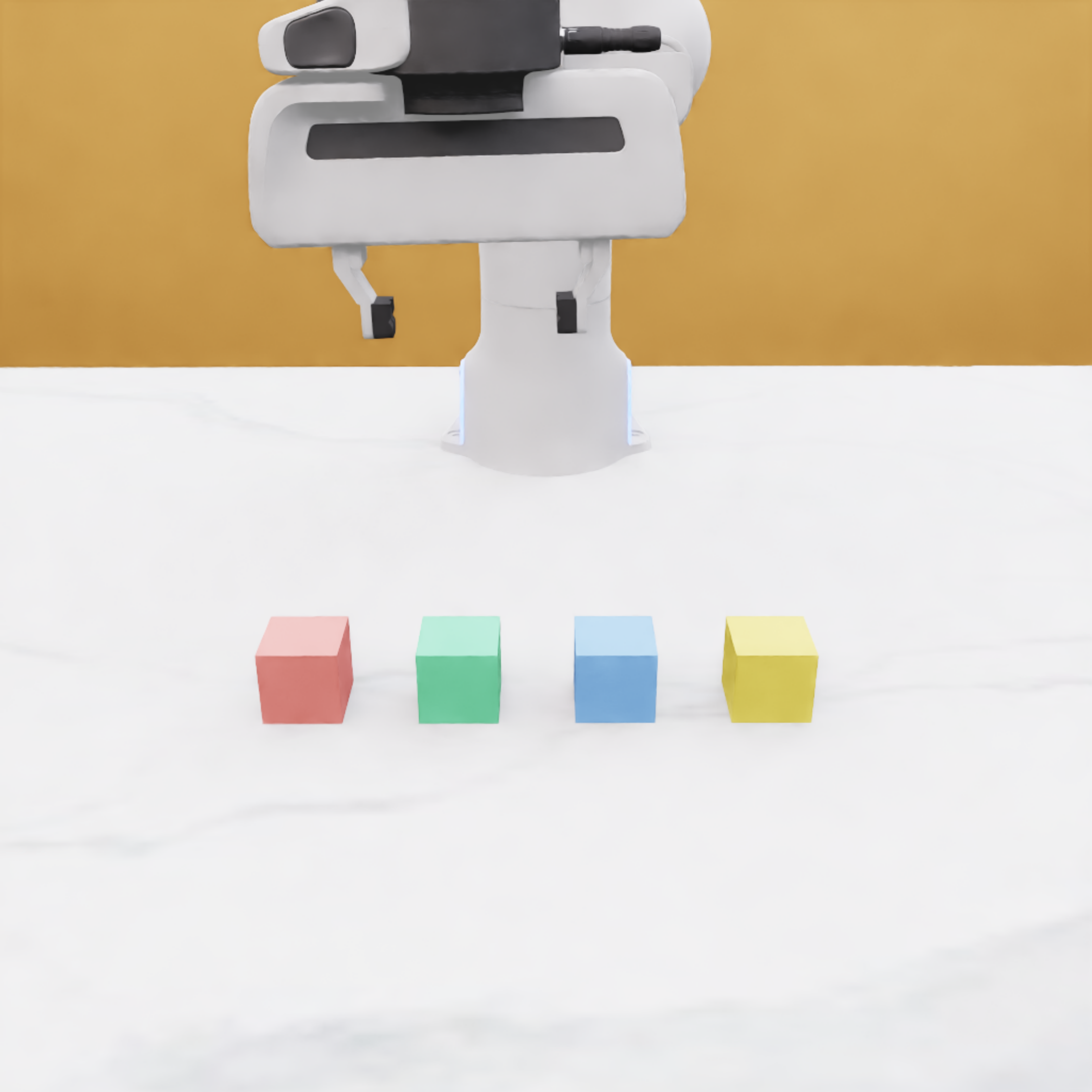}
{Move the end effector near the leftmost cube after reordering.}
{runtime\_block\_line structure.}
{Protocol target object reindex\_block\_1 proximity region.}
{$\mathbf{p}^*=(0.42, -0.05, 0.03)$ m; $\theta^*=0.00$ rad.}
{Success requires the end-effector distance to the target object's center to be $< 0.06$ m and the final state to remain stable for 1.0 s.}

\geomindtaskcard
{ReindexAugVisual\_ReindexExtreme\_v0}
{\geomindaug}
{Scene Updating}
{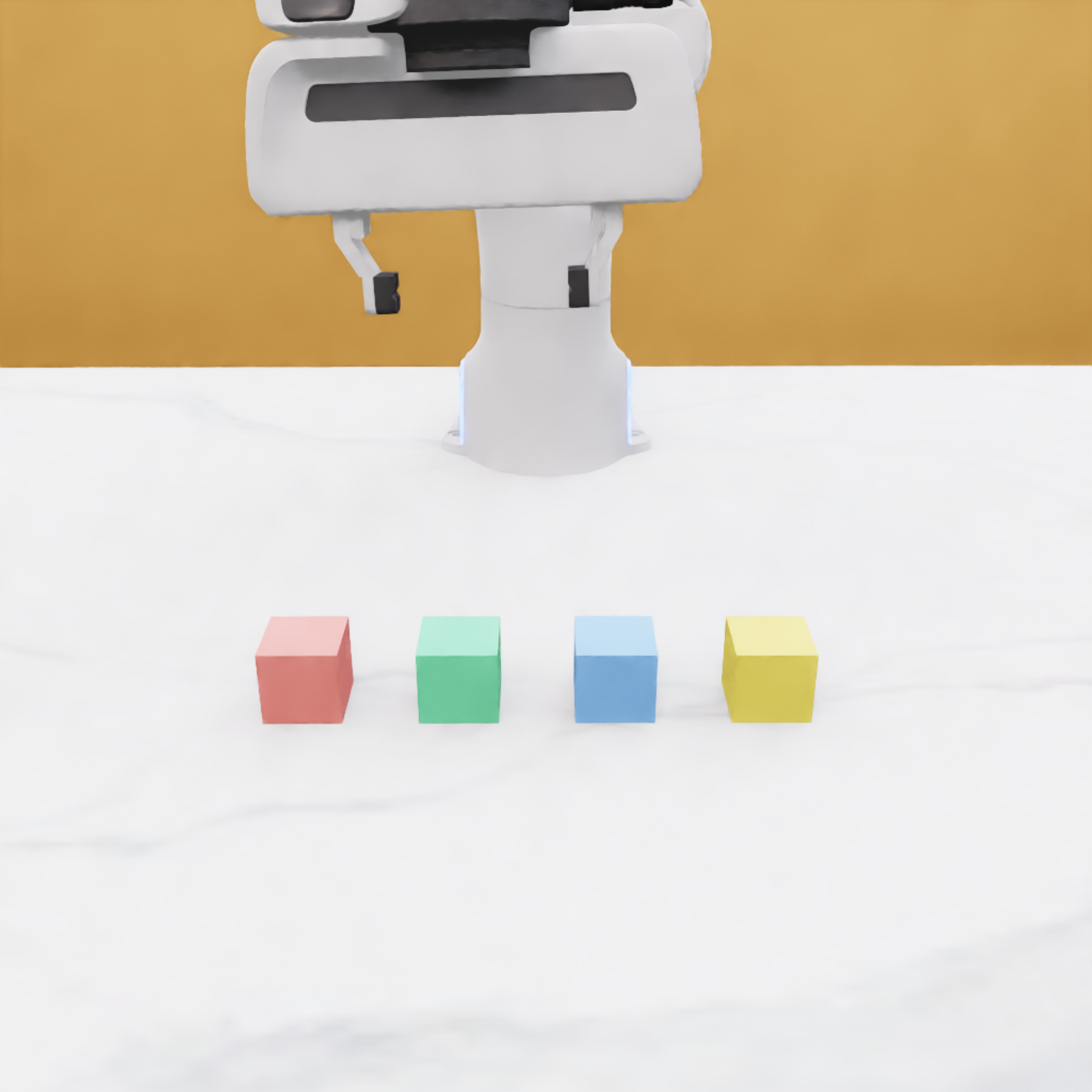}
{Move the end effector near the leftmost cube after reordering.}
{runtime\_block\_line structure.}
{Protocol target object reindex\_block\_1 proximity region.}
{$\mathbf{p}^*=(0.42, -0.05, 0.03)$ m; $\theta^*=0.00$ rad.}
{Success requires the end-effector distance to the target object's center to be $< 0.06$ m and the final state to remain stable for 1.0 s.}

\geomindtaskcard
{RelationAugSem\_PlaceDirectional\_v0}
{\geomindaug}
{Relational Grounding}
{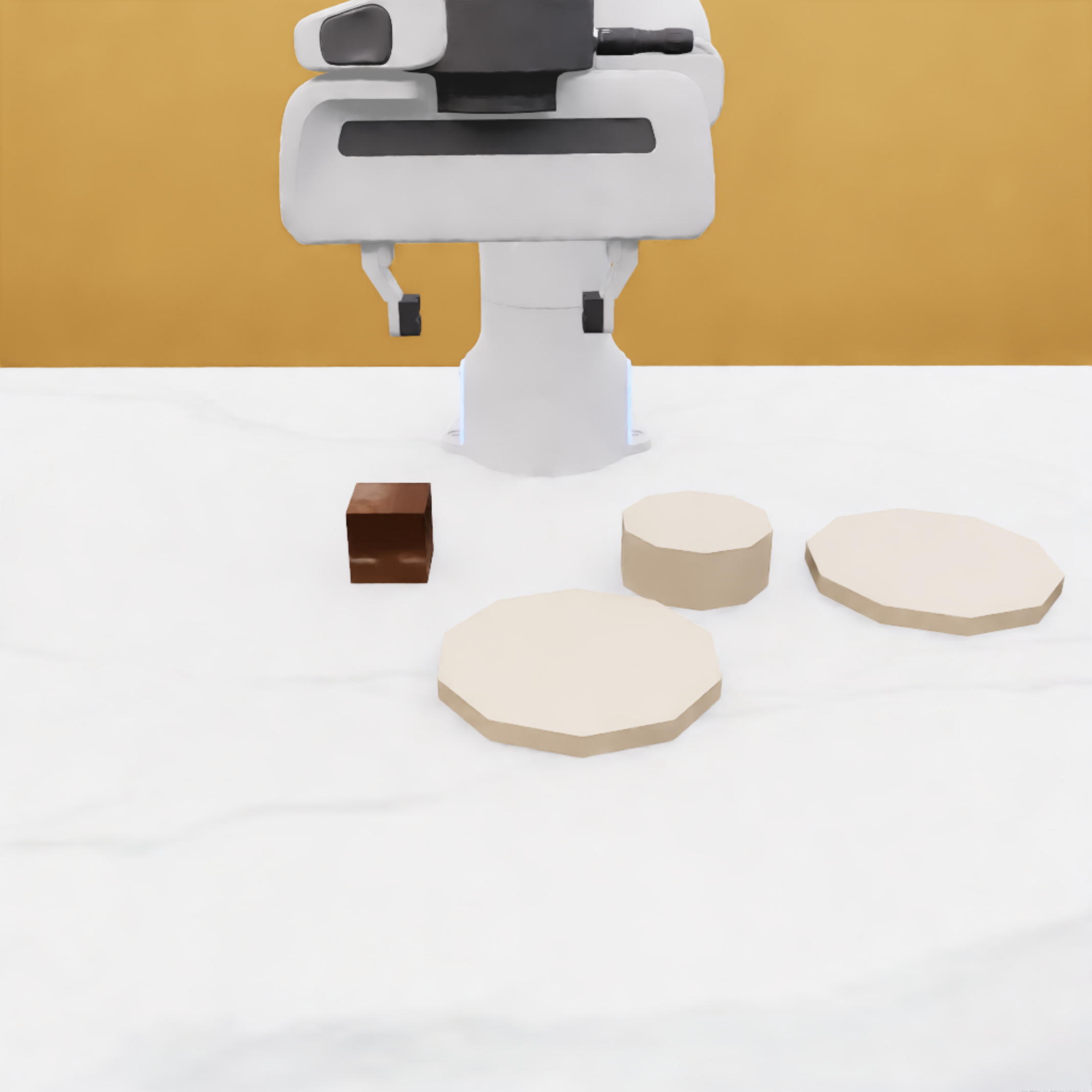}
{You already hold the wooden cube. Place it in the valid target region in front of the blue plate.}
{directional\_region\_semantic\_distractors structure.}
{Directional target zone relative to blue\_plate.}
{$\mathbf{p}^*=(0.57, 0.02, 0.03)$ m; $\theta^*=0.00$ rad.}
{Success requires the distance from the object center to the target position to be $< 0.09$ m and the final state to remain stable for 1.0 s.}

\geomindtaskcard
{RelationAugVisual\_PlaceDirectional\_v0}
{\geomindaug}
{Relational Grounding}
{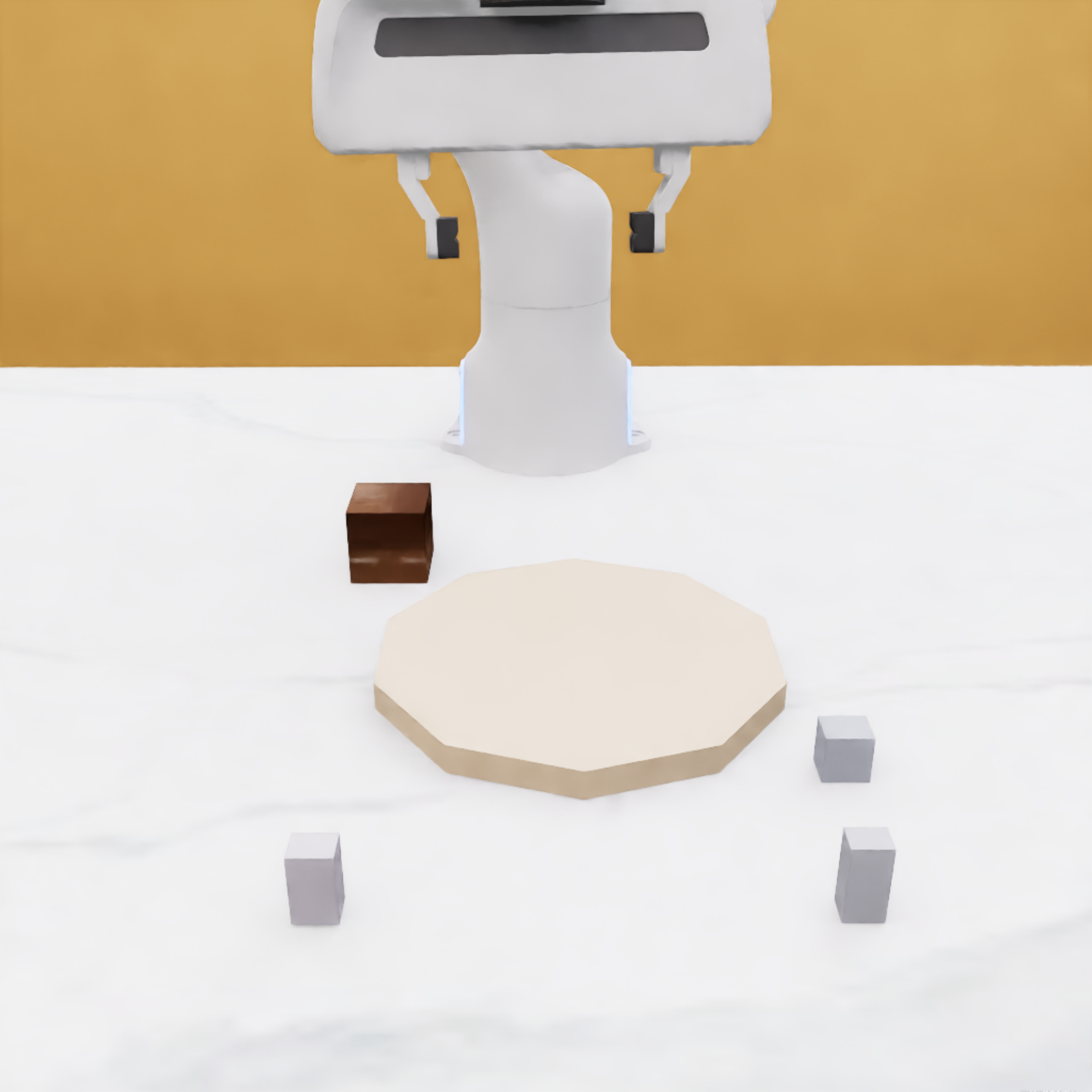}
{You already hold the wooden cube. Place it in the valid target region in front of the blue plate.}
{directional\_region\_visual\_clutter structure.}
{Directional target zone relative to blue\_plate.}
{$\mathbf{p}^*=(0.57, 0.02, 0.03)$ m; $\theta^*=0.00$ rad.}
{Success requires the distance from the object center to the target position to be $< 0.09$ m and the final state to remain stable for 1.0 s.}

\geomindtaskcard
{RelationBase\_PlaceDirectional\_v0}
{\geomindbase}
{Relational Grounding}
{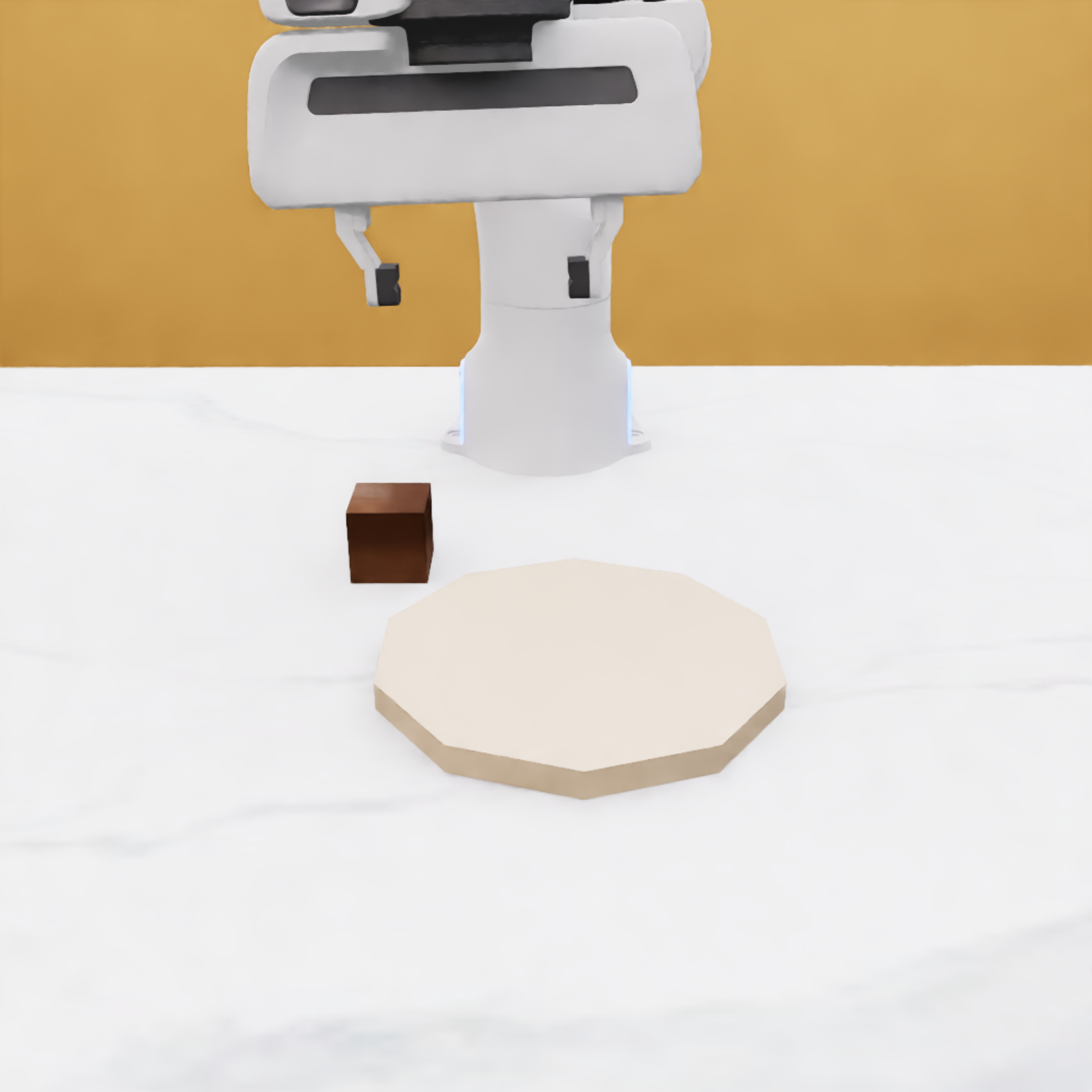}
{You already hold the wooden cube. Place it in the valid target region in front of the blue plate.}
{directional\_region\_injected structure.}
{Directional target zone relative to blue\_plate.}
{$\mathbf{p}^*=(0.65, 0.02, 0.03)$ m; $\theta^*=0.00$ rad.}
{Success requires the distance from the object center to the target position to be $< 0.09$ m and the final state to remain stable for 1.0 s.}

\geomindtaskcard
{RelationBase\_PlaceInside\_v0}
{\geomindbase}
{Relational Grounding}
{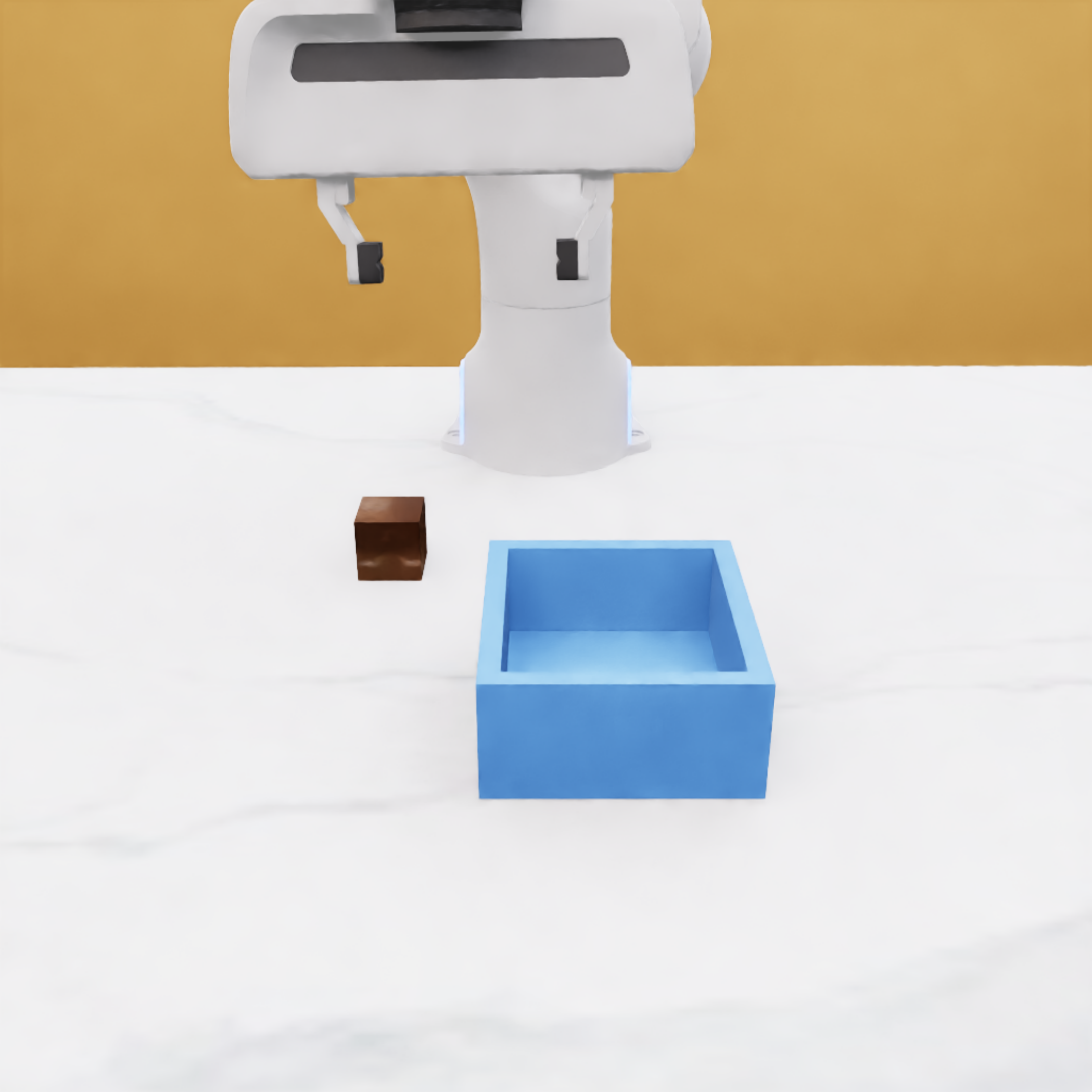}
{You already hold the cube. Put it fully inside the blue box.}
{containment\_standard\_injected structure.}
{Stable containment inside blue\_box\_floor.}
{$\mathbf{p}^*=(0.43, 0.04, 0.03)$ m; $\theta^*=0.00$ rad.}
{Success requires stable containment in the box, a center error $< 0.06$ m relative to the target position, and stability for 1.0 s.}

\geomindtaskcard
{RelationBase\_PlaceOnTop\_v0}
{\geomindbase}
{Relational Grounding}
{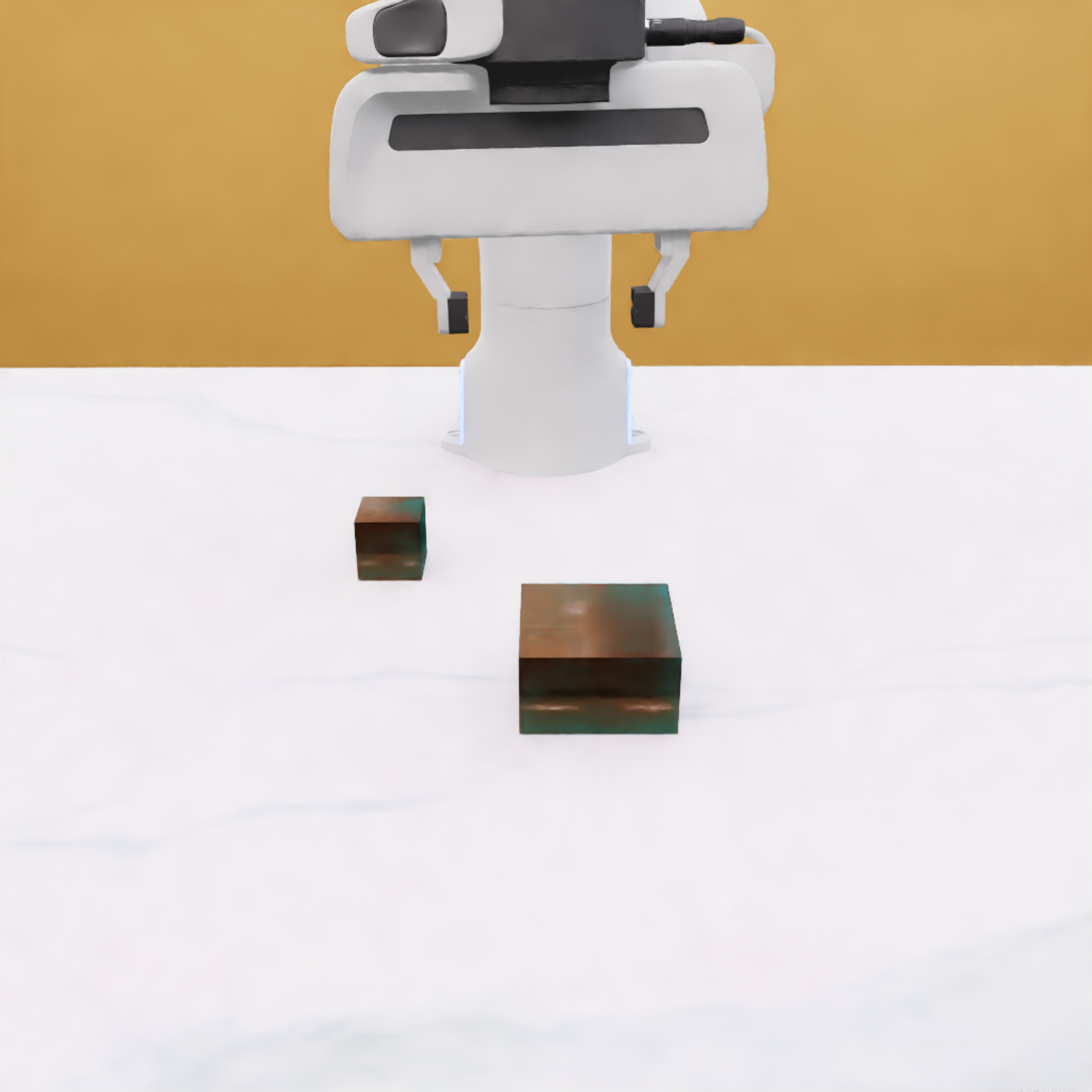}
{You already hold the cube. Release it onto the top surface of the red wooden block so it rests stably.}
{support\_top\_injected structure.}
{Stable top placement on red\_support\_block.}
{$\mathbf{p}^*=(0.41, 0.03, 0.07)$ m; $\theta^*=0.00$ rad.}
{Success requires stable placement on the support surface, a placement error $< 0.01$ m, and stability for 1.0 s.}

\geomindtaskcard
{TrackBase\_TrackClutter\_v0}
{\geomindbase}
{Identity Tracking}
{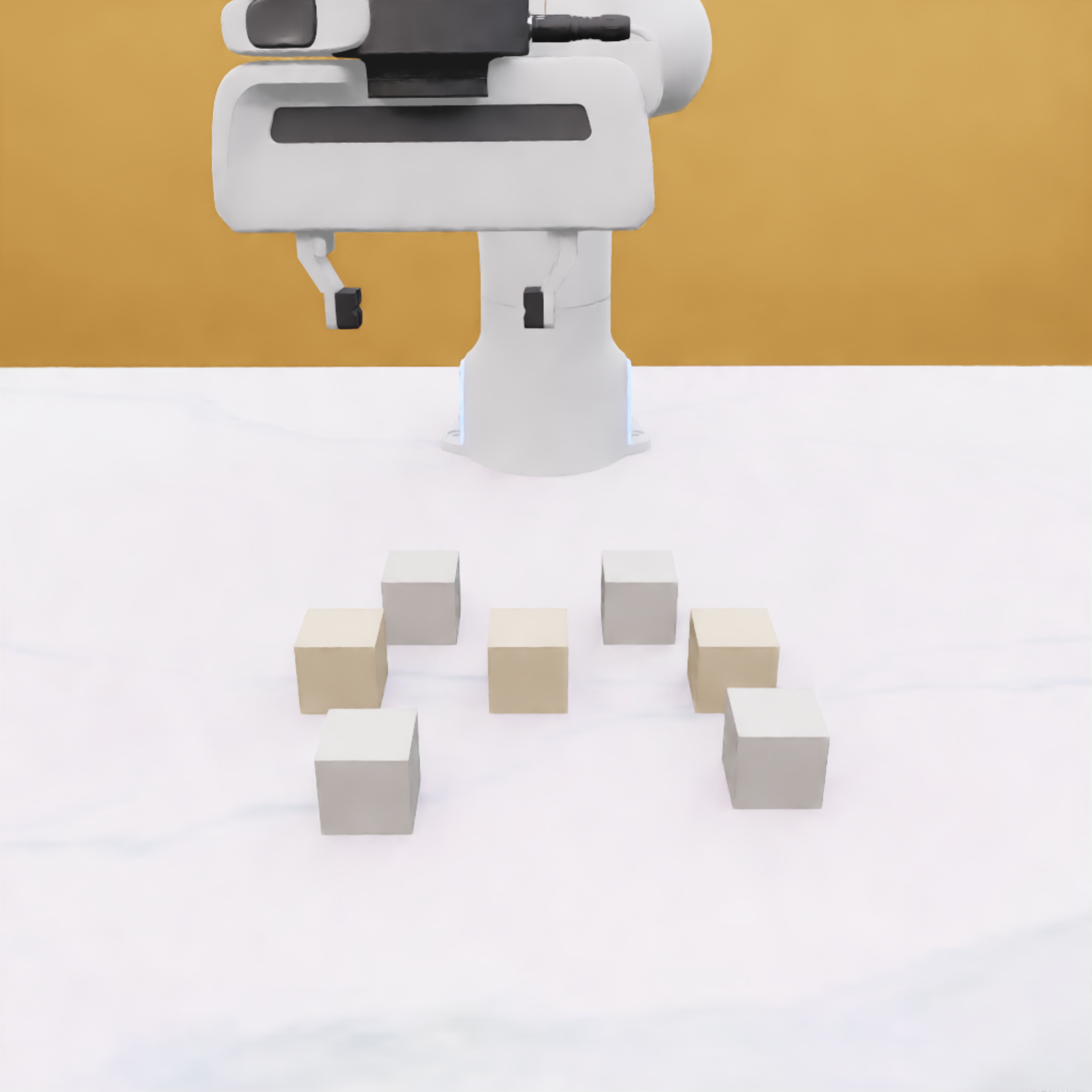}
{After the clutter settles, move the end effector near the cube that was initially in the middle.}
{clutter\_robust\_runtime\_identity\_line structure.}
{Protocol target object identity\_block\_1 proximity region.}
{$\mathbf{p}^*=(0.41, -0.01, 0.03)$ m; $\theta^*=0.00$ rad.}
{Success requires the end-effector distance to the target object's center to be $< 0.06$ m and the final state to remain stable for 1.0 s.}

\geomindtaskcard
{TrackBase\_TrackMulti\_v0}
{\geomindbase}
{Identity Tracking}
{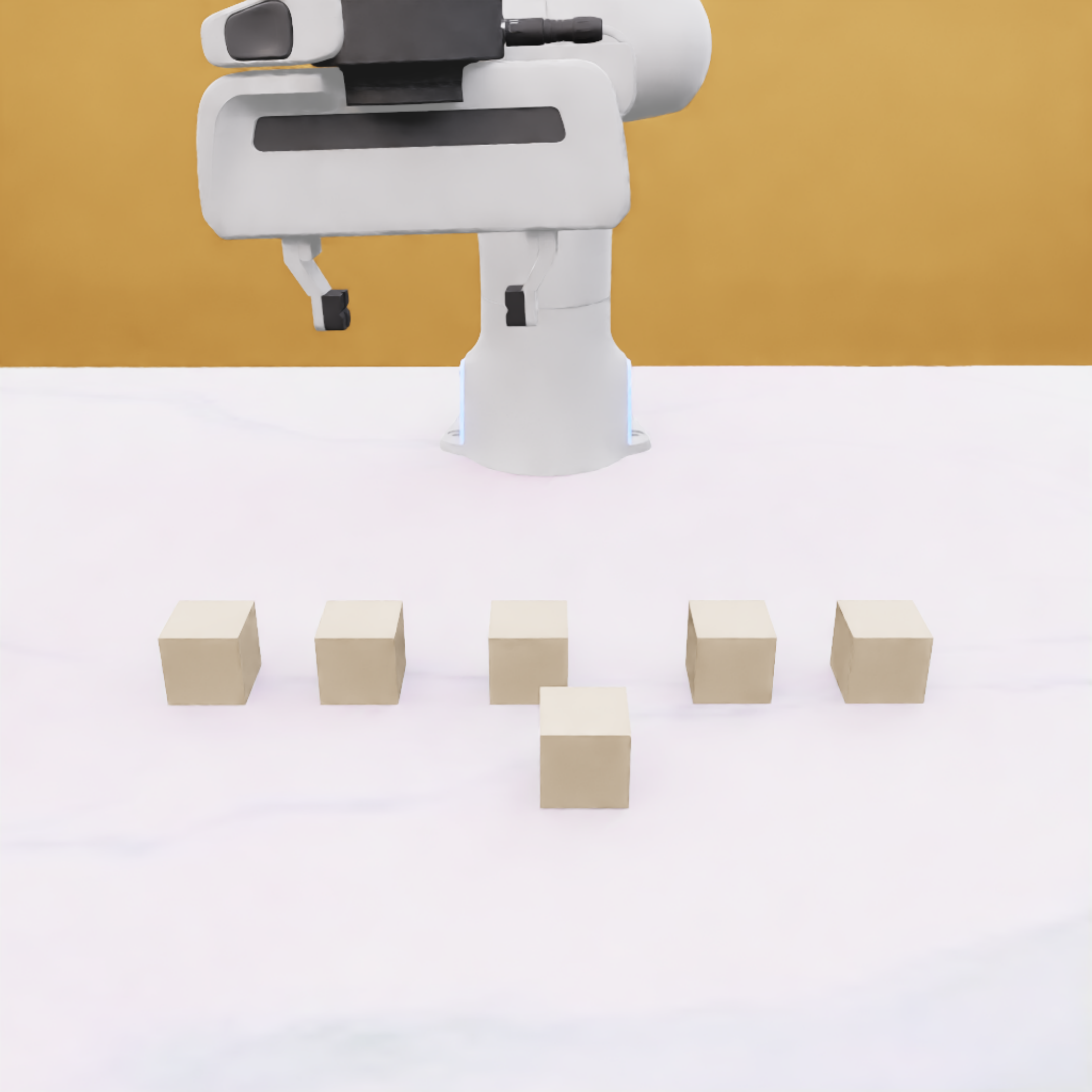}
{Move the end effector near the cube that started at the far left.}
{runtime\_identity\_multistep\_line structure.}
{Protocol target object identity\_block\_0 proximity region.}
{$\mathbf{p}^*=(0.40, -0.20, 0.03)$ m; $\theta^*=0.00$ rad.}
{Success requires the end-effector distance to the target object's center to be $< 0.06$ m and the final state to remain stable for 1.0 s.}

\geomindtaskcard
{TrackBase\_TrackSingle\_v0}
{\geomindbase}
{Identity Tracking}
{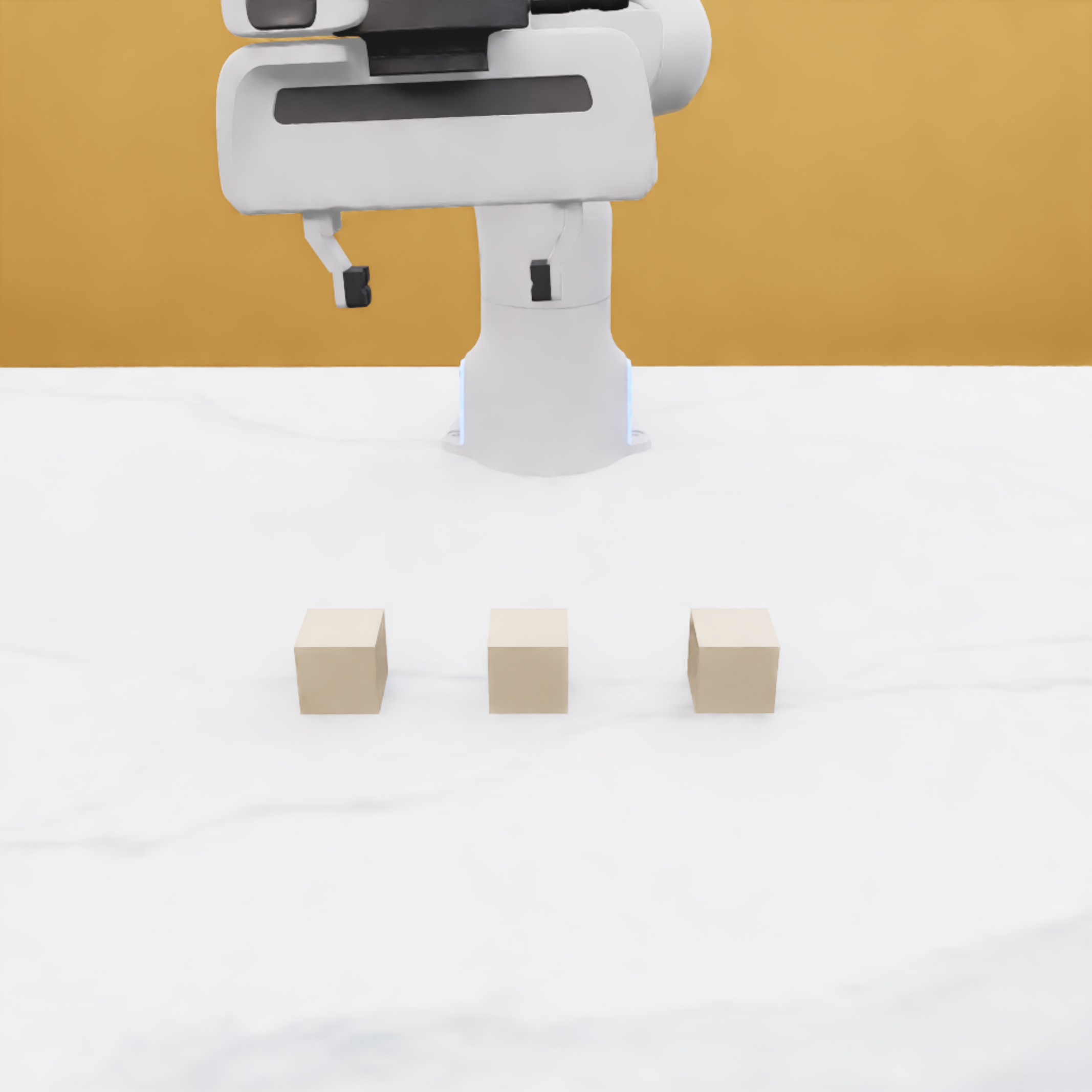}
{Move the end effector near the cube that started in the middle.}
{runtime\_identity\_line structure.}
{Protocol target object identity\_block\_1 proximity region.}
{$\mathbf{p}^*=(0.41, -0.01, 0.03)$ m; $\theta^*=0.00$ rad.}
{Success requires the end-effector distance to the target object's center to be $< 0.06$ m and the final state to remain stable for 1.0 s.}

\geomindtaskcard
{TrackAugSem\_TrackSingle\_v0}
{\geomindaug}
{Identity Tracking}
{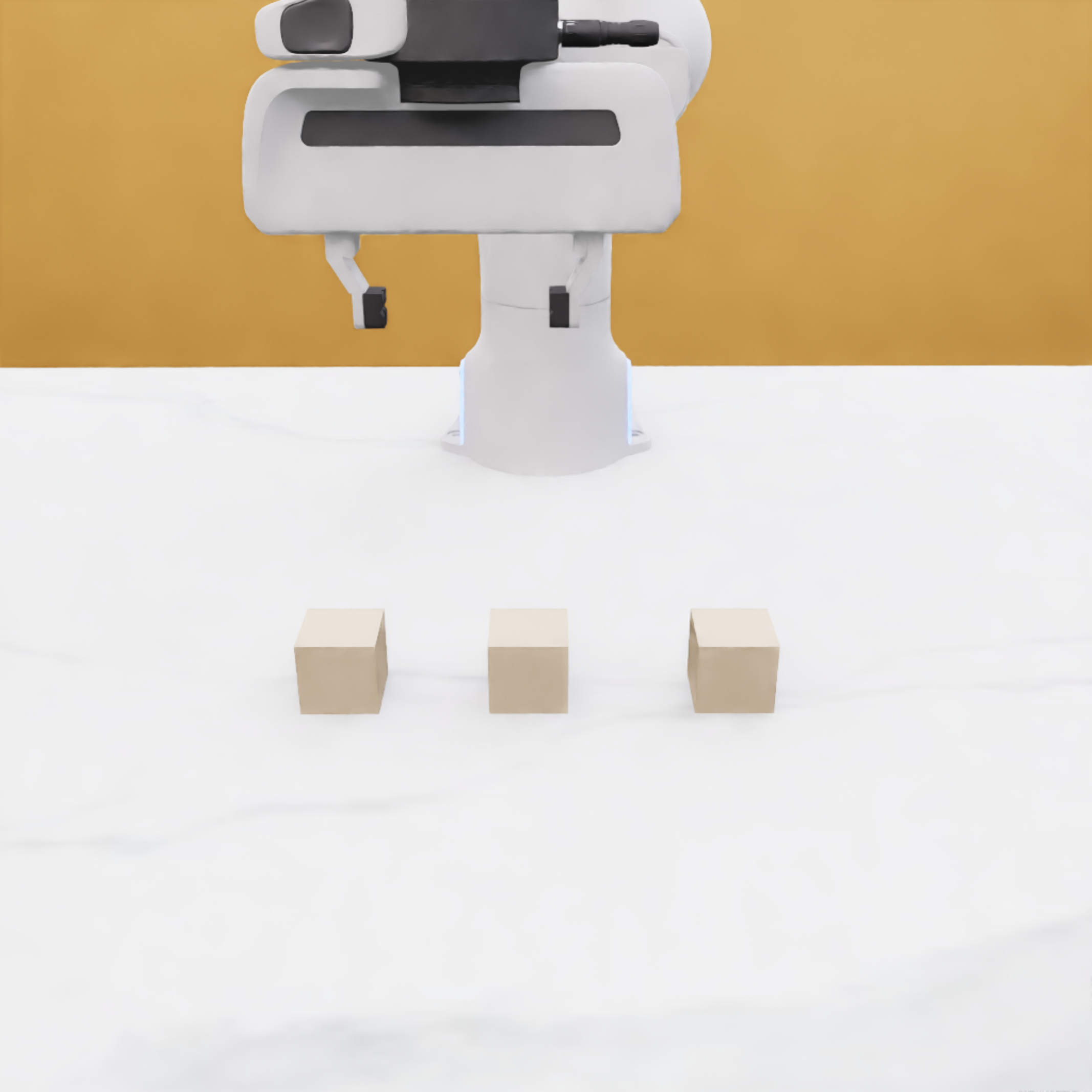}
{Move the end effector near the cube that started in the middle.}
{runtime\_identity\_line structure.}
{Protocol target object identity\_block\_1 proximity region.}
{$\mathbf{p}^*=(0.41, -0.01, 0.03)$ m; $\theta^*=0.00$ rad.}
{Success requires the end-effector distance to the target object's center to be $< 0.06$ m and the final state to remain stable for 1.0 s.}

\geomindtaskcard
{TrackAugVisual\_TrackSingle\_v0}
{\geomindaug}
{Identity Tracking}
{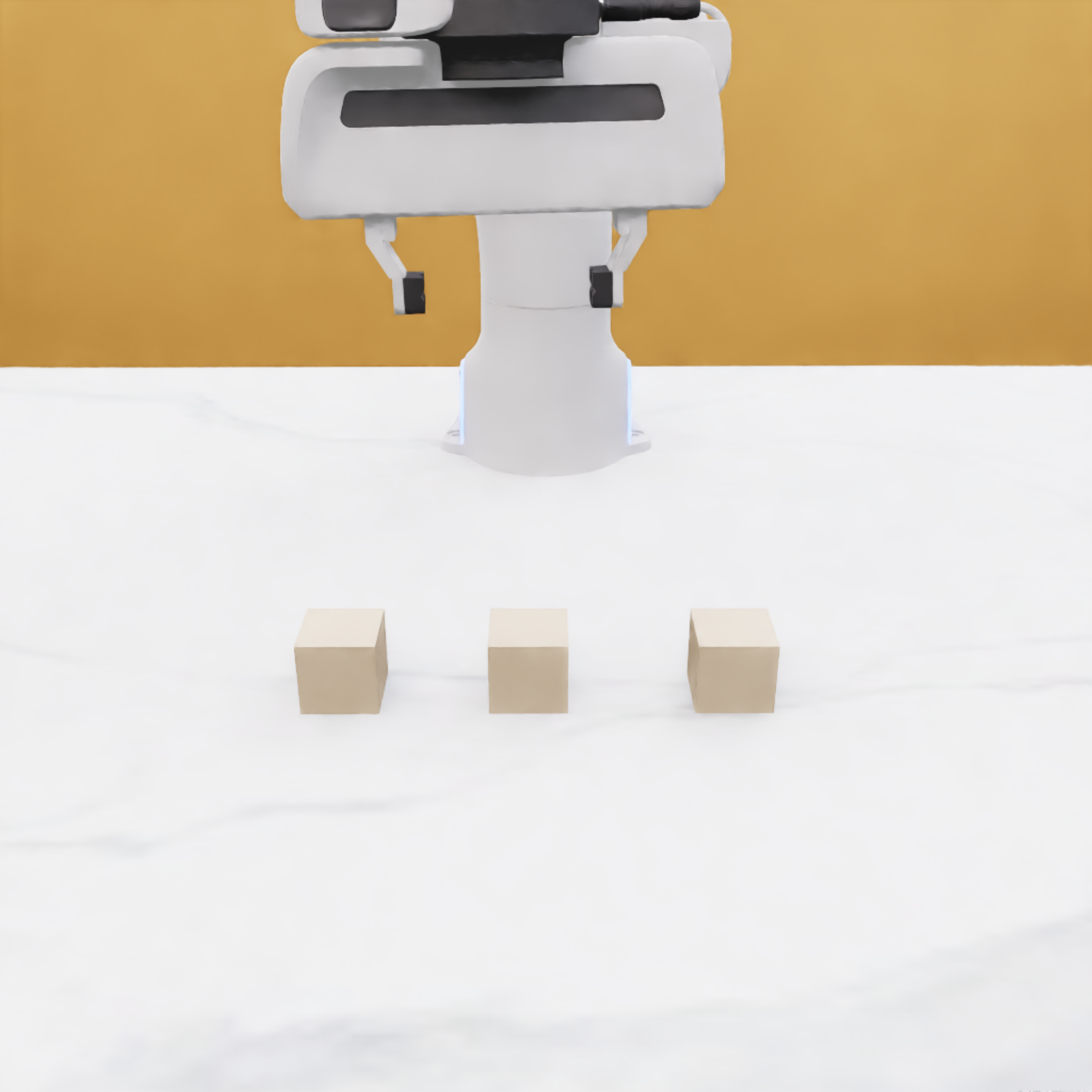}
{Move the end effector near the cube that started in the middle.}
{runtime\_identity\_line structure.}
{Protocol target object identity\_block\_1 proximity region.}
{$\mathbf{p}^*=(0.41, -0.01, 0.03)$ m; $\theta^*=0.00$ rad.}
{Success requires the end-effector distance to the target object's center to be $< 0.06$ m and the final state to remain stable for 1.0 s.}


\end{document}